\documentclass{article}

     \usepackage[preprint]{neurips_2025}

\usepackage[utf8]{inputenc} 
\usepackage[T1]{fontenc}    
\usepackage{hyperref}       
\usepackage{url}            
\usepackage{booktabs}       
\usepackage{amsfonts}       
\usepackage{nicefrac}       
\usepackage{microtype}      
\usepackage{xcolor}         
\usepackage{graphicx}
\usepackage{subfigure}
\usepackage{amsmath}
\usepackage{amssymb}
\usepackage{mathtools}
\usepackage{amsthm}
\usepackage[capitalize,noabbrev]{cleveref}

\usepackage{amsmath,amsfonts,bm}

\def\eqref#1{equation~\ref{#1}}

\def\1{\bm{1}}

 \def\d{\mathrm{d}}

\def\vzero{{\bm{0}}}

\def\vmu{{\bm{\mu}}}

\def\vepsilon{{\bm{\epsilon}}}

\def\va{{\bm{a}}}

\def\vh{{\bm{h}}}

\def\vt{{\bm{t}}}

\def\vw{{\bm{w}}}
\def\vx{{\bm{x}}}
\def\vy{{\bm{y}}}
\def\vz{{\bm{z}}}

\def\mI{{\bm{I}}}

\def\mR{{\bm{R}}}

\DeclareMathAlphabet{\mathsfit}{\encodingdefault}{\sfdefault}{m}{sl}
\SetMathAlphabet{\mathsfit}{bold}{\encodingdefault}{\sfdefault}{bx}{n}

\def\gN{{\mathcal{N}}}

\newcommand{\E}{\mathbb{E}}

\newcommand{\R}{\mathbb{R}}

\usepackage{amsmath, amssymb, amsfonts}
\usepackage{booktabs}
\usepackage{mathtools}
\usepackage {makecell} 
\usepackage{multirow}
\usepackage{algpseudocode}
\usepackage{pifont} 
\usepackage[textsize=tiny]{todonotes}
\theoremstyle{plain}
\newtheorem{theorem}{Theorem}[section]
\newtheorem{proposition}[theorem]{Proposition}
\newtheorem{lemma}[theorem]{Lemma}

\theoremstyle{definition}

\theoremstyle{remark}

\usepackage[ruled,vlined]{algorithm2e}

\usepackage{subcaption}
\usepackage{float} 

\let\cite\citep

\title{Straight-Line Diffusion Model for Efficient 3D Molecular Generation}

\author{\normalsize
  Yuyan Ni\textsuperscript{\textmd{1,2,3}}\thanks{Equal contribution. \textsuperscript{\dag} Work was done during Yuyan's internship at AIR. \textsuperscript{\ddag} Corresponding author: Yanyan Lan (\texttt{lanyanyan@air.tsinghua.edu.cn}).}  \ \textsuperscript{\dag} 
  \quad Shikun Feng\textsuperscript{\textmd{2} *} \quad 
  Haohan Chi\textsuperscript{\textmd{2}}\quad 
Bowen Zheng\textsuperscript{\textmd{4}}\quad 
Huan-ang Gao\textsuperscript{\textmd{2}}\\
\normalsize
\textbf{Wei-Ying Ma}\textsuperscript{\textmd{2}}\quad 
\textbf{Zhi-Ming Ma}\textsuperscript{\textmd{1}}\quad
\textbf{Yanyan Lan}\textsuperscript{\textmd{2,5} \ddag}\\
 \textsuperscript{\textmd{1}} Academy of Mathematics and Systems Science, Chinese Academy of Sciences \\
 \textsuperscript{\textmd{2}} Institute for AI Industry Research (AIR), Tsinghua University 
 \\
 \textsuperscript{\textmd{3}} University of Chinese Academy of Sciences
 \\
 \textsuperscript{\textmd{4}}Huazhong University of Science and Technology
 \\
 \textsuperscript{\textmd{5}}Beijing Academy of Artificial Intelligence
}

\begin{document}



\maketitle

\vspace{-0.5cm}
\begin{abstract}
Diffusion-based models have shown great promise in molecular generation but often require a large number of sampling steps to generate valid samples. In this paper, we introduce a novel Straight-Line Diffusion Model (SLDM) to tackle this problem, by formulating the diffusion process to follow a linear trajectory. The proposed process aligns well with the noise sensitivity characteristic of molecular structures and uniformly distributes reconstruction effort across the generative process, thus enhancing learning efficiency and efficacy. Consequently, SLDM achieves state-of-the-art performance on 3D molecule generation benchmarks, delivering a 100-fold improvement in sampling efficiency.\footnote[1]{The code is open-sourced at  \url{https://github.com/fengshikun/SLDM}}
\end{abstract}
\vspace{-0.5cm}
\section{Introduction}

3D molecular generation is an essential task in drug discovery, material science, and molecular engineering. The goal is to computationally design 3D molecular structures that not only capture intricate physical and chemical constraints but also fulfill specific properties.

\begin{figure}[h!]
\vskip-0.05in
  \begin{minipage}{0.53\textwidth}
        Recently, diffusion models have been widely applied in this field, inspired by their remarkable success in image synthesis \cite{dhariwal2021diffusion,rombach2022stablediffusion,peebles2023scalabledit}, and other domains \cite{brooks2024video,abramson2024accurate}. Methods like EDM \cite{hoogeboom2022equivariant}, EDM-Bridge \citep{wu2022diffusion} and GeoLDM \cite{xu2023geometric} have demonstrated the potential of diffusion-based frameworks to generate chemically valid 3D molecular structures. However, these direct applications of diffusion methods usually require a large number of sampling steps to produce valid molecules. Taking EDM as an example, it requires approximately 1000 steps of function evaluations to generate molecules with around 82$\%$ stability, which is a key metric for assessing sample quality by quantitatively measuring whether the molecule satisfies chemical constraints. Reducing the sampling steps significantly degrades the molecule stability.
  \end{minipage}
  \hfill
  \begin{minipage}{0.45\textwidth}
    \includegraphics[width=1\textwidth]{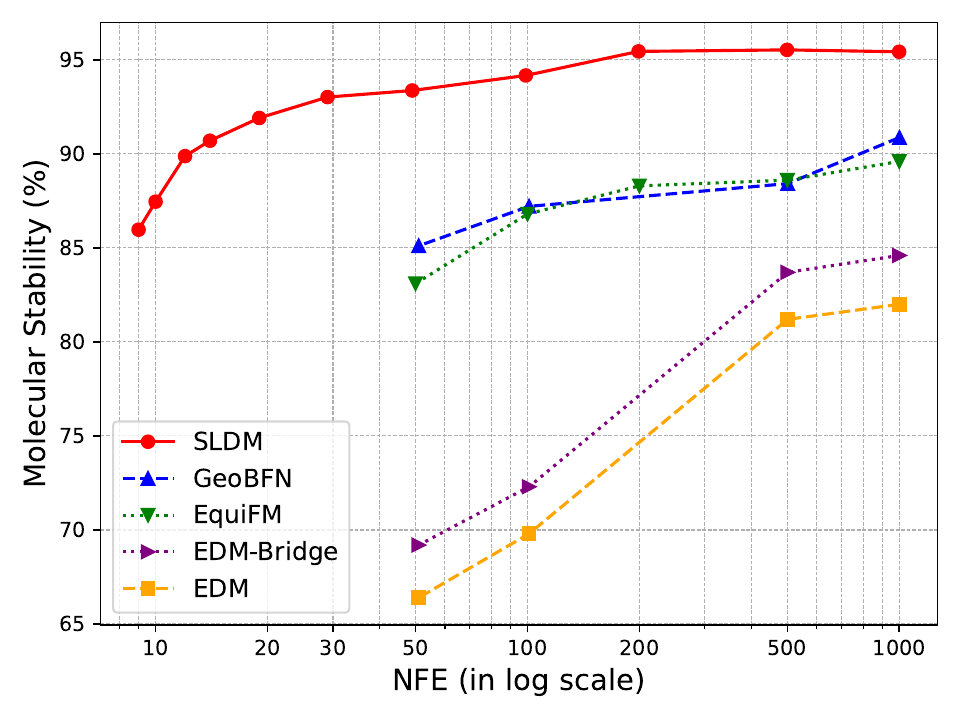}
    \vskip -0.1in
    \caption{Comparison of molecule stability ($\uparrow$) across diffusion-based molecular generation models on QM9 unconditional generation, evaluated by the number of function evaluations (NFE) during the sampling process. }
    \label{fig:efficiency}
  \end{minipage}
  \vskip-0.1in
\end{figure}

To improve sampling efficiency, EquiFM \cite{song2024equiFM} and GeoBFN \cite{song2023geobfn} have been proposed to utilize the Flow Matching (FM) framework \cite{tong2023conditionalflowmatching} and Bayesian Flow Networks (BFN) \cite{graves2023bayesian} for molecular generation. The use of these advanced generative AI models enables a speedup of 5$\times$ and 20$\times$, respectively, compared to EDM. However, they still require a large number of sampling steps (e.g.~1000) to achieve high molecule stability (e.g.~90$\%$), as shown in Figure \ref{fig:efficiency}.

To understand why existing methods suffer from low efficiency, we analyze the issue through the lens of truncation error in sampling. We begin by establishing a unified perspective on previous diffusion-based methods, including diffusion models, flow matching methods, and Bayesian flow networks. Specifically, their noise corrupting process can be generalized as $ \vx_t=\mu(t)\vx_0+\sigma(t)\vepsilon, \vepsilon\sim \mathcal{N}(\vzero,\mI_N)$, where $\vx_0$ represents the clean data, $\vx_t$ is the noise corrupted data at time $t\in[0,T]$, $\mu(t)$ and $\sigma(t)$ define the schedule of the process. In this framework, all these processes can be equivalently framed as continuous-time Ordinary Differential Equations (ODEs), even though they may employ stochastic sampling in practice. This viewpoint allows the sampling process to be interpreted as a numerical approximation of the solution trajectory of the underlying ODE. Crucially, most existing methods rely on first-order estimation, whose truncation error is governed by the second-order term $\frac{\d^2 x(t)}{\d t^2}(\Delta t)^2$. We observe that $\frac{\d^2 x(t)}{\d t^2}$ in these approaches can be large, requiring small step sizes $\Delta t$ to reduce the truncation error, which results in a large number of sampling steps.

To address this issue, we propose a novel diffusion process called Straight-Line Diffusion Model (SLDM). The key idea is to minimize truncation error by striving to achieve a linear sampling trajectory, i.e.~$\frac{\d^2 x(t)}{\d t^2}=0$. This approach allows the diffusion dynamics to tolerate larger step sizes without sacrificing accuracy, leading to a substantial improvement in sampling efficiency. Building on this objective, we theoretically prove that when $\mu(t)=1-t/T$ and $\sigma$ is a small constant, the process guarantees a near-linear trajectory. Intuitively, this process features a linearly decreasing mean term and a consistently small variance term, representing a smooth linear progression from the origin point to the data.

Notably, SLDM strikes a good balance between efficiency and efficacy by using our straight-line schedule. Firstly, this strategy aligns well with the inductive bias of molecular generation, preventing the introduction of chemically implausible conformations. Unlike images, molecular structures are much more sensitive to noise, and even small perturbations can lead to unrealistic structures that violate chemical principles. This challenge requires a slower signal-to-noise ratio (SNR) decay during the noise-adding process, as suggested in \citet{song2023geobfn}. By naturally satisfying a slower SNR decay, our method maintains chemical information of the intermediate states, enhancing computational efficiency compared to traditional methods such as EDM. Secondly, our strategy achieves a more balanced generative process, significantly improving the model’s learning efficacy. In methods like GeoBFN, minimal perturbations are applied in the later stages, shifting most of the reconstruction burden to earlier stages. Although this reduces the computational load in the later stages, it creates an uneven distribution of effort, limiting the model's learning capacity. In contrast, our approach evenly distributes the reconstruction effort across the entire process, enabling the model to learn effectively at each stage. This balance results in a more stable and efficient learning process, enhancing the robustness and accuracy of the generated molecular structures.

We conduct extensive experiments to demonstrate the potential of straight-line diffusion in 3D molecular generation and other domains. As shown in Figure \ref{fig:efficiency}, using only at most 10 or 15 sampling steps, SLDM surpasses EDM or EquiFM, GeoBFN with 1000 sampling steps, achieving up to \textbf{100- or 70-fold} improvement in sampling efficiency. In terms of generation quality, SLDM with 200 sampling steps achieves 95\% molecular stability, significantly outperforming the best baseline, GeoBFN, which requires 1000 steps to reach 90\% molecular stability. We also observe that similar improvements can be achieved when applying SLDM to the conditional generation task, i.e. generating molecules with a desired property, highlighting its potential to enable more practical and controllable molecular design in future applications.

\section{Analysis on Sampling Efficiency}\label{sec:unify diffusion}


We begin by theoretically analyzing the underlying factors contributing to the sampling efficiency issue. In particular, we first present a unified framework for diffusion-based methods, including their SDE and ODE formulations. We then examine the sampling truncation error from the ODE perspective, highlighting the critical role of the process's second-order derivative in improving sampling efficiency.

We denote the data as $\vx\in\R^N$, where for molecules, it refers to a 3D point cloud comprising atomic coordinates and potentially other atomic features. According to \citet{karras2022elucidating,xue2024unifyingbfn}, various diffusion-based models, including DDPM~\citep{ho2020denoising}, DDIM~\citep{songjiam2020denoising}, VE~\citep{song2021scorebased}, 
FM~\citep{lipman2023flow}, and BFN~\citep{graves2023bayesian}, can be formulated as a unified form with the noise corrupting process defined as:
\begin{equation}\label{eq:xt def}
    \vx_t=\mu(t)\vx_0+\sigma(t)\vepsilon, \vepsilon\sim \mathcal{N}(\vzero,\mI_N), t\in[0,T]
\end{equation} where $\vx_0,\vx_t$ are clean data and noise corrupted data respectively, $\mu(t)$ and $\sigma(t)$ define the schedule of the process. Specifically, as detailed in Appendix \ref{sec:app summarize}, the schedule parameters are summarized as follows: $\mu_{\text{DDPM(EDM)}}\!=\!1\!-\!(t/T)^2$, $\mu_{\text{VE}}\!=\!\mu_{\text{DDIM}}\!=\!1$, $\mu_{\text{BFN}}=1-\sigma_{min}^{2(1-t/T)}$, $\mu_{\text{FM}}\!=\!1\!-\!t/T$; $\sigma_{\text{DDPM(EDM)}}\!=\!\sqrt{1\!-\!(1\!-\!(t/T)^2)^2}$, $\sigma_{\text{VE}}\!=\!\sqrt{t}$, $\sigma_{\text{DDIM}}\!=\!{t}$, $\sigma_{\text{BFN}}\!=\!\sqrt{\mu_{\text{BFN}}(1\!-\!\mu_{\text{BFN}})}$, $\sigma_{\text{FM}}\!=\!t/T\!+\!(1\!-\!t/T)\sigma_{min}$, where DDPM(EDM) uses the approximated DDPM schedule given in EDM \cite{hoogeboom2022equivariant}. $\sigma_{min}$ are defined as small constants to ensure $\mu(0)\approx 1$ and $\sigma(0)\approx 0$. $T$ is typically chosen to be sufficiently large so that $\vx_T$ approximates a known distribution.


Extending a similar theoretical technique from \citet{karras2022elucidating} to the unified form, we can prove that \eqref{eq:xt def} is the solution to the following linear stochastic differential equation (SDE):
\begin{equation}\label{eq:sde}
    \d\vx_t =\frac{\dot{\mu}(t)}{\mu(t)}\vx_t \d t+\sqrt{2\sigma(t)\dot{\sigma}(t)-2\sigma(t)^2\frac{\dot{\mu}(t)}{\mu(t)}}\d w_t,
\end{equation}
where $\mu(t)$ and $\sigma(t)$ should adhere to the constraints $\mu(0)=1 $ , $\sigma(0)=0$, with $\mu(t)\geq 0 $ being monotone non-increasing, $\sigma(t)\geq 0$, and $\sigma(t)/\mu(t)$ being monotone non-decreasing. The detailed proof is elaborated in Appendix~\ref{secapp:unify formulation}. 

According to the ODE and SDE relations revealed in \citet{song2021scorebased}, we can derive the equivalent ODE that follows the same marginal probability densities as the above SDE:  
\begin{equation}\label{eq:ODE}
    \d\vx_t \!=\!\left[\!\frac{\dot{\mu}(t)}{\mu(t)}\vx_t\!-\!\!\left(\!\sigma(t)\dot{\sigma}(t)\!-\!\sigma(t)^2\frac{\dot{\mu}(t)}{\mu(t)}\!\right) \!\!\nabla_\vx\!\log p_t(\vx_t)\!\right]\!\d t.
\end{equation}

Consequently, most existing diffusion-based methods widely used in 3D molecular generation, including DDPM, DDIM, VE, FM, and BFN, can be interpreted from a continuous-time perspective, where their diffusion process can equivalently be viewed as an SDE in \eqref{eq:sde} or an ODE in \eqref{eq:ODE}. Therefore, we can use the ODE formulation as a valuable perspective to analyze the sampling efficiency issue. 

Specifically, the diffusion sampling process can be interpreted as a numerical approximation of the backward solution trajectory of the underlying ODE in \eqref{eq:ODE} \footnote{Connections between sampling of baseline methods and first-order ODE sampling are discussed in appendix \ref{sec:app ode sample}.}. This numerical approximation inherently introduces truncation errors at each step. For example, under Euler's method, the simplest and widely-used numerical scheme, $x(t-\Delta t)$ is approximated by $x(t)- \frac{\d x(t)}{\d t}\Delta t$ when solving the ODE backward in time. However, the true value can be derived from the Taylor expansion:
\begin{equation}
    x(t-\Delta t)=x(t)- \frac{\d x(t)}{\d t}\Delta t+\frac{1}{2}\frac{\d^2 x(t)}{\d t^2}(\Delta t)^2+O((\Delta t)^3),
\end{equation} where $\Delta t$ is a small step size. Thus, the truncation error primarily arises from the second-order term, governed by $\frac{\d^2 x(t)}{\d t^2}$. 

To minimize this truncation error, traditional numerical analysis primarily focuses on developing higher-order solvers to estimate higher-order terms, assuming the ODE structure is fixed. EquiFM \cite{song2024equiFM} applied such a technique in 3D molecular generation to speedup the sampling process. However, higher-order solvers face a trade-off between the number of function evaluations (NFE) and accuracy, limiting their ability to achieve even smaller NFEs. As a result, the issue remains partially unsolved.
In contrast, we take a different approach: rather than focusing on the sampling algorithm, we reformulate the diffusion process itself to minimize the truncation error, which directly influences the sampling efficiency.


However, our unified formulation of diffusion introduces a key flexibility: the ability to modify the schedule of the process, thereby altering the ODE structure itself to reduce these errors. Therefore, we emphasize that the key to minimizing the truncation error is to reduce the second-order derivative of the process.

\section{Straight-Line Diffusion Process}

To reduce the truncation error of the sampling process, we aim to design a new diffusion process whose inherent ODE exhibits minimal $\frac{\d^2 x(t)}{\d t^2}$. By achieving this, even a basic Euler iteration can deliver low truncation error without resorting to complex solvers that require multiple function evaluations. This novel perspective advances the Pareto frontier of efficiency and accuracy in diffusion sampling.

\subsection{Derivation of SLDM}\label{sec:straight line}

Given the intractability of the score function for general data distributions, we begin by examining a simplified case where the initial distribution is a delta distribution. This setup provides a tractable backward ODE, enabling a more straightforward analysis. In this case, the only solution that ensures \textcolor{black}{$\frac{\d\vx}{\d t}$} remains constant is for $\sigma(t)$ to be constant and $\mu(t)$ to be a linear function of $t$. Details can be found in Appendix \ref{app sec: straight line}.

Given the boundary condition that $x_0$ approximates data distribution and $x_{t=1}$ approximates a known distribution, the only feasible choice is to set $\sigma(t)$ to a small poistive constant and $\mu(t)=1-t/T$, which satisfies all the schedule constraints noted under \eqref{eq:sde}, except $\sigma(0)=0$. This is because setting $\sigma(0) = 0$ exactly would result in a trivial denoising objective, To avoid this, we instead approximate $\sigma(0)$ by a small value, which keeps the denoising task non-trivial while still satisfying the boundary condition approximately. In practice, we use a value for $\sigma$ that is two orders of magnitude smaller than the data scale, yielding good results.

Rescaling time to the interval $[0,1]$, the resulting diffusion process is:
\begin{equation}\label{eq:str schedule}
    \vx_t=(1-t)\vx_0+\sigma\vepsilon, t\in[0,1],
\end{equation}
which ensures a straight-line trajectory under the delta data distribution assumption.

We then analyze the above diffusion process for general data distribution and demonstrate that a near-linear trajectory could be achieved by setting a small constant value for $\sigma$, as shown in the following theorem.
\begin{theorem}[Near-linear Trajectory of SLDM]\label{thm: linear_trajectory}
    For a general data distribution and the schedule in \eqref{eq:str schedule}, the following inequality holds for each dimension $i$:
    \begin{equation}
        P(|\frac{\d\vx_t^{(i)}}{\d t}+\frac{\vx_t^{(i)}}{1-t}|\geq \delta)\leq\frac{\sigma^2}{\delta^2(1-t)^2}.
    \end{equation}
When $\sigma\to 0$, $\frac{\d\vx_t}{\d t}+\frac{\vx_t}{1-t}$ converges to zero in probability, where the solution to equation $\frac{\d\vx_t}{\d t}+\frac{\vx_t}{1-t}=0$ is that $\frac{\vx_t}{1-t}$ is constant, which corresponds to a linear trajectory.
\end{theorem}
\textcolor{black}{In other words, for $t\in[0,1-\frac{\sigma}{\delta\cdot \epsilon}]$, $|\frac{\d\vx_t^{(i)}}{\d t}+\frac{\vx_t^{(i)}}{1-t}|< \delta$ holds with probability at least $1-\epsilon^2$, indicating that setting $\sigma$ to a small value ensures that the trajectory remains close to a straight line for most timesteps. This aligns with our empirical observations shown in Figure \ref{fig:compare schedule trajectory}: the overall SLDM trajectory is straighter compared to other diffusion processes, and the sampling trajectory deviates more from a straight line in the initial steps but becomes increasingly linear later. This initial deviation introduces sampling error, but empirically, we find that this error can be mitigated by the Langevin dynamics component in our stochastic sampler introduced in Section~\ref{sec:sampling}.}

\begin{figure*}[t]
	\centering
     \includegraphics[width=1\linewidth]{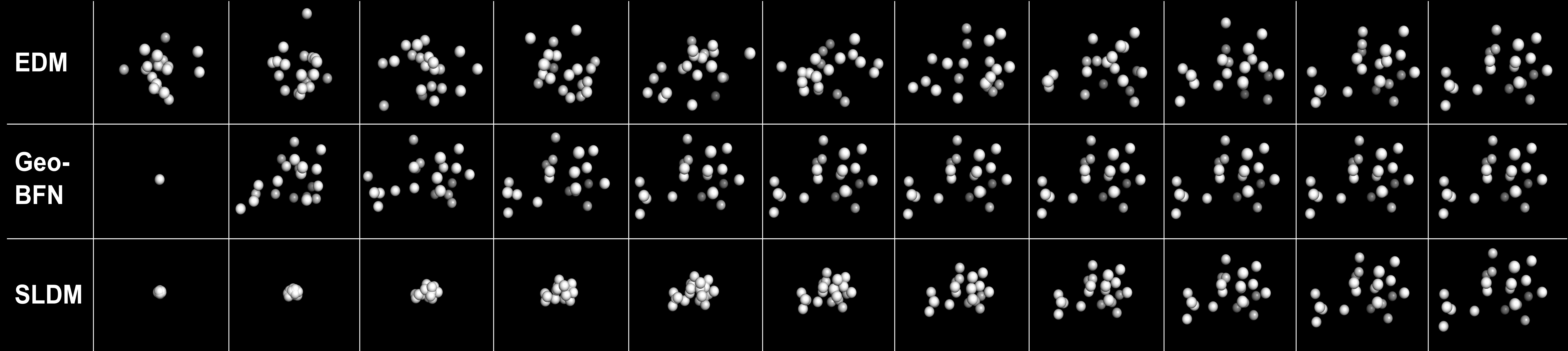}
     \vskip -0.03in
    \caption{The diffusion process of atomic coordinates in EDM, GeoBFN and SLDM. }
    \label{fig:compare mol process}
    \vskip -0.1in
\end{figure*}

 \subsection{Comparisons with Previous Diffusion Processes}\label{sec:SNR}
Previous studies \cite{nichol2021improved,rombach2022stablediffusion} have underscored the importance of amortizing the generative difficulty through the diffusion process to improve sample quality. These works typically customize signal-to-noise ratio (SNR) schedules to adapt to varying data characteristics. Notably, as pointed out in \citet{song2023geobfn}, the point cloud representation of molecular structures is far more sensitive to noise compared to images. Consequently, for molecular generation, the SNR needs to be decreased at a much slower pace than in the image generation domain.
\begin{figure}[h]
  \centering
  \begin{minipage}{0.45\textwidth}       
        Our proposed method aligns remarkably well with these insights. As shown in Figure~\ref{fig:compare schedule mu sigma}, the SNR in our method decreases significantly slower than that of previous methods. 
         In addition, our schedule results in a smoother and more stable generative process for molecules, as demonstrated in Figure~\ref{fig:compare mol process}. The process unfolds uniformly from the origin, preserving the relative spatial relationships of atomic coordinates in intermediate states and retaining critical chemical information throughout the generative trajectory.

         In contrast, traditional diffusion models like EDM, which follows the process schedule of DDPM, also known as VP, involve a noise 
  \end{minipage}
  \hfill
  \begin{minipage}{0.53\textwidth}
     \includegraphics[width=1\linewidth]{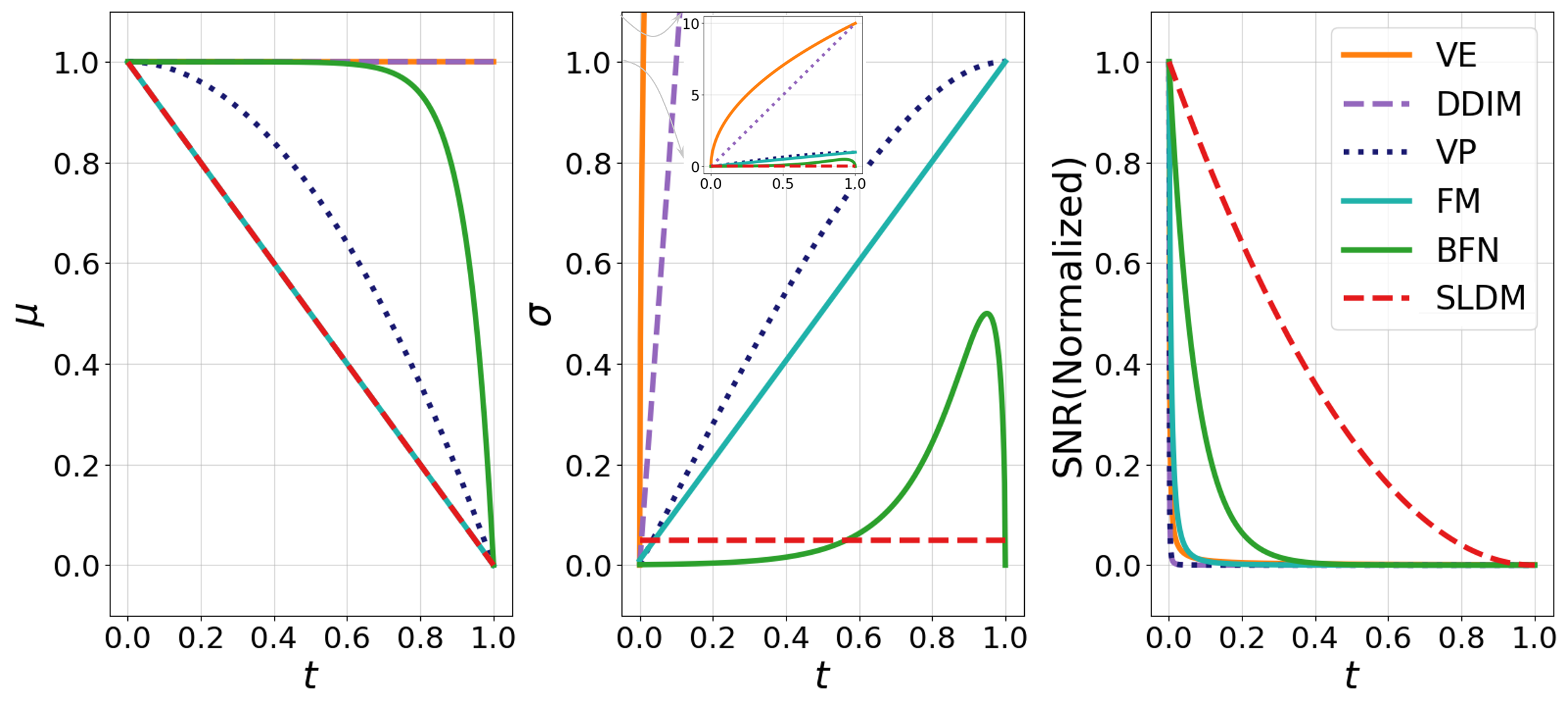}	
    \caption{Comparison of schedule parameters across diffusion-based models. Here SNR refers to $(\mu/\sigma)^2$. \textcolor{black}{For SLDM, we set $\sigma = 0.05$, consistent with experiments. The cutoff values $\sigma_{\text{max}}$ and $\sigma_{\text{min}}$ used in other baselines are provided in Appendix~\ref{sec:app summarize}.} }
    \label{fig:compare schedule mu sigma}
\end{minipage}
\vskip-0.26in
\end{figure}

variance that increases monotonically during the forward process. This increasing noise quickly disrupts the spatial structure of molecules, rendering much of the diffusion process chemically meaningless. As a result, many computational resources are wasted on steps that attempt to reconstruct signal-less states, leading to a severe reduction in overall model efficiency. Though GeoBFN adopts a low-noise regime for most timesteps, it keeps $\mu\approx 1$ and $\sigma\approx 0$ for over a half of the process, making little changes to the molecular structure. Therefore, GeoBFN shifts the majority of reconstruction difficulty to the earlier stages, creating a severely unbalanced generative process. 


To sum up, compared to existing diffusion-based models, the new schedule in SLDM aligns better with the inductive biases of molecular data. The approach of distributing the reconstruction difficulty more evenly across the entire diffusion process helps facilitate more effective learning, improving the model’s efficiency and efficacy.

 


\subsection{Sampling Strategy of SLDM}\label{sec:sampling}
Here we introduce the specific sampling strategy of SLDM. First, we need to derive the reverse process of \eqref{eq:sde} or equivalently \eqref{eq:ODE}. Notably, previous work \cite{xue2024sa} (Equation 6) provides a unified form of the reverse-time stochastic differential equation (SDE), which shares the same marginal distribution as the forward process, thereby ensuring that the sampling process has the ability to reconstruct the data distribution. We apply the parameter $\mu$ and $\sigma$ of SLDM to this equation and then discretize it using the Euler-Maruyama method, yielding the following discretized reverse-time SDE:
\begin{equation}\label{eq:ode+LD}
\begin{aligned}    
    \vx_{t-\Delta t}
   = \underbrace{\vx_t+ \frac{\Delta t}{1-t}\left(\vx_t+\sigma^2 \nabla_\vx \log p_t(\vx_t)\right)}_{\text{ODE Sampling of \eqref{eq:ODE}} } \underbrace{+\beta(t)\frac{\Delta t}{1-t}\sigma^2 \nabla_\vx \log p_t(\vx_t)\!+\!\sqrt{2\beta(t)\frac{\Delta t}{1-t}} \sigma\vepsilon}_{\text{Langevin dynamics}},
\end{aligned}
\end{equation}
where $\beta(t)$ is any non-negative bounded function. 

Similar to previous works \cite{xue2024sa,karras2022elucidating}, this sampling can be interpreted as a combination of ODE sampling and Langevin dynamics. The ODE component drives the denoising process along deterministic trajectories, while the Langevin dynamics introduce stochastic corrections. Specifically, we choose $\beta(t)=(1-t)/\Delta t$, to ensure the expectation of the RHS of \eqref{eq:ode+LD} is $\E[\vx_{t-\Delta t}|\vx_t] $, as it provides the optimal estimation of $\vx_{t-\Delta t}$ in terms of minimizing the mean squared error (MSE). Details are demonstrated in Appendix \ref{secapp:unbiased sampling}.  

Thus the sampling algorithm becomes:
\begin{equation}\label{eq:our sampling}
\begin{aligned}   
   \vx_{t-\Delta t}=&\frac{1-(t-\Delta t)}{1-t}(\vx_t\!+\!\sigma^2 \nabla_\vx \log p_t(\vx_t))\!+\!\sqrt{2}\sigma\vepsilon.
\end{aligned}
\end{equation} 
Note that $1-t$ appears in the denominator of the above equation. To avoid potential division by zero, we choose to skip the \(t=1\) sampling step and find that it works well. This might be because the early sampling steps are less critical to the final result, likely due to Langevin dynamics not strictly requiring a specific prior, which imparts an inherent error-correction capability to the algorithm. This property is further evidenced by our observation that the sampling results are relatively robust to the initial distribution's variance.

\begin{figure}[t]
\begin{minipage}[t]{0.4\textwidth}
\begin{algorithm}[H]
\caption{SLDM Training}
   \label{alg:training}
\KwIn{data $\vx_0$ with dimension $N$, neural network $\phi$, variance constant $\sigma$}
\Repeat{$converged$}{
    \hspace{0.2em} Sample $t \sim U([0,1])$\\
    Sample $\vepsilon\sim \gN(\textbf{0}, I_N)$\\
    Compute $\vx_t = (1-t)\vx_0 + \sigma\vepsilon$\\
    Minimize $|| \vepsilon-\phi(\vx_t, t)||^2$   
}
\KwRet{$\phi$}
\end{algorithm}
\end{minipage}
\hfill
\begin{minipage}[t]{0.56\textwidth}
\begin{algorithm}[H]
\caption{SLDM Sampling}
   \label{alg:sampling}
\KwIn{trained network $\phi$, same $N$ and $\sigma$ as training, sampling steps $T$, temperature annealing rate $\nu$}
\hspace{0.2em}Sample $\vx \sim \sigma\cdot\gN(\textbf{0}, I_N)$\\
\For{$i = T-1$ \textbf{to} $1$}{
    $\vx\gets\frac{T-(i-1)}{T-i}\cdot(\vx-\sigma \cdot \phi(\vx,t))$\\
   \If{$i>1$}{
    $\vx\gets\vx+{(\frac{i}{T})}^{\nu}\sqrt{2}\cdot\sigma\cdot\vepsilon$, $\vepsilon\sim \gN(\textbf{0}, I_N)$    }
}
\KwRet{ $\vx$}
\end{algorithm}
\end{minipage}
\vskip-0.2in
\end{figure}




The above sampling algorithm provides a principled way to approximately sample from the input data distribution, but its practical application in molecular datasets presents unique challenges. In particular, the molecular stability of the dataset is not guaranteed to be 100\%. To enhance stability and mitigate the impact of data noise, low-temperature sampling can be employed to generate samples that preserve essential properties of the input data.
For Langevin dynamics, the sampling temperature can be controlled by scaling the stochastic term with a constant. However, prior work has demonstrated that such temperature control in diffusion model sampling often fails to effectively balance diversity and fidelity \cite{dhariwal2021diffusion}. Additionally, as observed in our toy dataset experiments, conventional low-temperature sampling suffers from mode collapse, failing to fully cover all high-density regions of the target distribution.

To address these limitations, we propose a time-annealing temperature schedule:
\begin{equation}\label{eq:sde-decay}
    \vx_{t-\Delta t}=\frac{1-(t-\Delta t)}{1-t}(\vx_t+ \sigma^2\nabla_\vx \log p_t(\vx_t))+  t^\nu\sqrt{2}\sigma\vepsilon,
\end{equation}
where \(\nu\) controls the decay rate of temperature over time. This approach allows for higher stochasticity during the initial stages of sampling, enabling the exploration of distant probability density maxima. As the temperature decreases in later stages, the process converges to the probability density maxima. Further theoretical explanations and toy data illustrations are provided in Appendix \ref{sec app temp}. 
With the help of the temperature control, our sampling strategy can reduce the impact of noise in the data, and generate molecules with enhanced stability.
The complete training and sampling procedure of straight-line diffusion are given in algorithm \ref{alg:training} and \ref{alg:sampling}. \textcolor{black}{We further discussed the relation to relevant techniques used in general generative approaches in Related Work (Appendix \ref{sec:discussion Relevant Techniques}), including noise scheduling, flow-based methods, other straight-line trajectory models, etc.}

\begin{table*}[h]
\caption{Unconditional molecular generation results on QM9 and GEOM-Drugs datasets. For all diffusion-based models, $T$ denotes sampling steps. Metrics are calculated with 10000 samples generated from each model. Higher values indicate better performance. }
\label{exp:main_gen}
\vskip-0.2in
\setlength{\tabcolsep}{2.5pt}
\centering
\begin{tabular}{ccccccc}
\\ \toprule
\multicolumn{1}{l|}{}          & \multicolumn{4}{c|}{QM9}                                                           & \multicolumn{2}{c}{GEOM-Drugs} \\ \hline
\multicolumn{1}{l|}{\#Metrics} & Atom sta(\%) & Mol sta(\%) & Valid(\%)   & \multicolumn{1}{c|}{V*U(\%)} & Atom sta(\%)   & Valid(\%)   \\
\hline
\multicolumn{1}{l|}{Data}       & 99.0 &95.2& 97.7  & \multicolumn{1}{c|}{97.7}       &  86.5 &99.9       \\
\hline
\multicolumn{1}{l|}{E-NF}       & 85.0 & 4.9    & 40.2        & \multicolumn{1}{c|}{39.4}       & -       & -        \\
\multicolumn{1}{l|}{G-Schnet}       & 95.7 & 68.1  & 85.5      & \multicolumn{1}{c|}{80.3}       & -      & -       \\
\multicolumn{1}{l|}{EDM (T=1000)}       & 98.7& 82.0        & 91.9         & \multicolumn{1}{c|}{90.7}       & 81.3       & 92.6        \\
\multicolumn{1}{l|}{GDM (T=1000)}       & 97.6 & 71.6        & 90.4        & \multicolumn{1}{c|}{89.5}       & 77.7       & 91.8     \\
\multicolumn{1}{l|}{EDM-Bridge (T=1000)}       & 98.8  & 84.6        & 92.0         & \multicolumn{1}{c|}{90.7}       & 82.4       & 92.8        \\
\multicolumn{1}{l|}{GeoLDM (T=1000)}       & 98.9  & 89.4        & 93.8         & \multicolumn{1}{c|}{92.7}       & 84.4       & 99.3        \\
\multicolumn{1}{l|}{EquiFM (T=200)}       & 98.9 & 88.3 & 94.7        & \multicolumn{1}{c|}{\underline{93.5}}       &  84.1 &98.9       \\
\multicolumn{1}{l|}{GeoBFN (T=1000)}      & 99.08 & 90.87 & 95.31&         \multicolumn{1}{c|}{92.96}       & 85.60 &92.08         \\
\multicolumn{1}{l|}{\textcolor{black}{END (T=1000)}}&98.9& 89.1& 94.8& \multicolumn{1}{c|}{92.6}&87.0& 89.2\\
\hline
\multicolumn{1}{l|}{SLDM (T=1000)}       &\textbf{99.43} &\textbf{95.42}&\textbf{97.07}& \multicolumn{1}{c|}{90.42}&\underline{88.30}&\textbf{99.95}\\
\multicolumn{1}{l|}{SLDM (T=50)}       & \underline{99.30}&\underline{93.37}&\underline{96.24}&\multicolumn{1}{c|}{\textbf{93.63}}&\textbf{89.03}&\underline{99.57}\\
\bottomrule
\end{tabular}
\end{table*}

\section{Experiment}
To validate the advantages of our method in molecular generation, we evaluate its overall performance and sampling efficiency in both unconditional and conditional generation scenarios.
\subsection{Setup}\label{sec:exp setup}
\textbf{Datasets}  We evaluate our model using two widely adopted datasets for unconditional molecular generation, with all dataset splitting strictly following baseline settings \citep{hoogeboom2022equivariant, song2024equiFM, song2023geobfn}.
QM9 \citep{ruddigkeit2012enumeration, ramakrishnan2014quantum} contains approximately 134,000 small organic molecules with up to nine heavy atoms. It is split into training (100K), validation (18K), and test (13K) sets.
GEOM-Drugs \citep{axelrod2022geom} focuses on drug-like molecules, comprising around 430,000 molecules with sizes ranging up to 181 atoms and an average of 44.4 atoms per molecule. Its larger size and greater diversity make it more challenging for generative models. The dataset is randomly divided into training, validation, and test sets using an 8:1:1 ratio.

For conditional molecular generation, we adopt the QM9 dataset with the same setup as prior work \citep{hoogeboom2022equivariant, song2024equiFM, song2023geobfn}. The QM9 training partition is split into two halves, each containing 50K samples. Specifically, the QM9 training set is divided into two halves of 50K samples each. The first half is used to train a classifier for ground-truth property labels, while the second half is used to train the conditional generative model.

\textbf{Implementation} 
The molecule is represented by atomic coordinates and atom types, \(\vz = (\vx, \vh)\), where \(\vx \in \R^{3M}\) denotes the atomic coordinates, $M$ is the number of atoms, and \(\vh\) encodes the atom type information. Thus for molecule generation, the model needs to generate both coordinates and atom types.
A key requirement for the molecular coordinates is ensuring the SE(3) invariance of the probability distribution, meaning that the probability of generating two molecular conformations should be identical if they only differ by translation or rotation. Following \citet{xu2022geodiff, hoogeboom2022equivariant}, we tailor the SLDM algorithm to satisfy equivariance. Specifically, to ensure translation invariance, we constrain the coordinates to a zero Center of Mass (CoM) space, while rotation invariance is preserved by employing equivariant neural networks to predict the noise. To ensure a fair comparison of generative algorithms, we use EGNN \citep{satorras2021n} as the backbone model, consistent with the baseline methods \citep{garcia2021n,hoogeboom2022equivariant,wu2022diffusion,xu2023geometric,song2024equiFM, song2023geobfn}. We prove that the generated data distribution satisfies SE(3) invariance, as provided in Appendix \ref{sec:app equivariance}.
For atom types, we follow  UniGEM \citep{feng2024unigem} to predict atom types based on the generated coordinates. The SLDM algorithms tailored for molecular generation are provided in Appendix~\ref{sec:algorithm}. Hyperparameters are summarized in Appendix~\ref{sec:app detail}. An introduction to the baseline models is included in Section \ref{sec: related work}.

\textcolor{black}{
\textbf{Baselines} To ensure a fair comparison, we select competitive baselines that also focus on generative modeling and have the same neural architectures (e.g., EGNN) and input information (e.g., coordinates and types). We also acknowledge that some studies propose strategies orthogonal to generative algorithms, as discussed in Related Work (Appendix \ref{sec: related work}). However, these strategies are not directly comparable to our work, as they do not aim to replace or improve the generative model itself but rather focus on improving network expressivity and augmenting input information. We believe that substituting their generative algorithms with ours could lead to improvements, and exploring this will be part of the future work.}

 \textcolor{black}{For conditional molecular generation, Two additional basic baselines: Random and $N_{\text{atoms}}$ follow EDM \citep{hoogeboom2022equivariant}. The Random baseline involves shuffling property labels in the training data and evaluating the property classifier on the shuffled data. The $N_{\text{atoms}}$ baseline uses the property classifier network to predict the property based solely on the number of atoms in the molecule.
}

\begin{figure*}[t]
	\centering
    \includegraphics[width=1\linewidth]{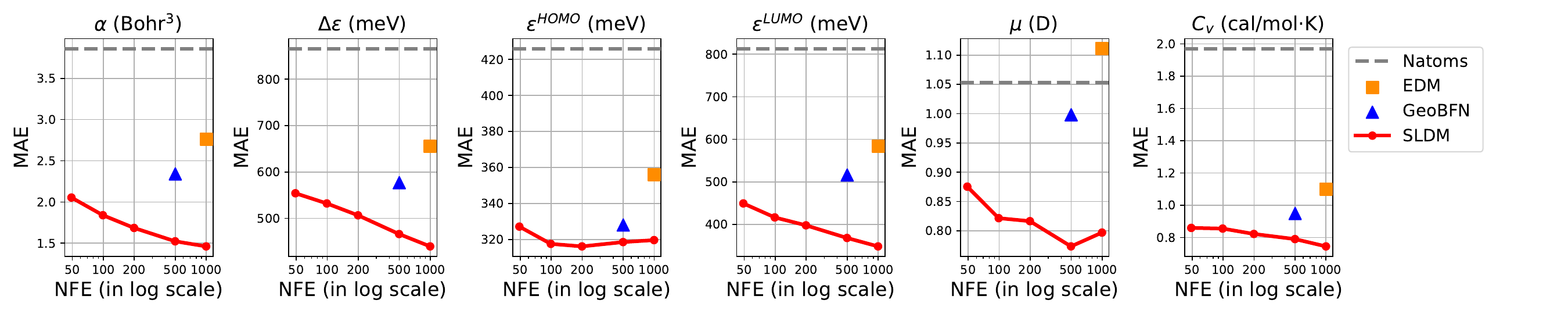}
\vskip -0.1in
    \caption{Comparison of performance in conditional generation ($\downarrow$), with respect to the number of function evaluations (NFE) during the sampling process. }
    \label{fig:cond efficiency}
    \vskip -0.1in
\end{figure*}

\textbf{Metrics} 
It is important to note that evaluation protocols differ across the literature, as discussed in \textcolor{black}{section} \ref{sec: related work}. To ensure consistency, we strictly adhere to the evaluation methods used in our baselines. We sample 10,000 molecules and 
predict the bond type (single, double, triple, or non-existent) based on the distances between each pair of atoms, as in \citet{hoogeboom2022equivariant}. Atom stability is computed as the proportion of atoms with correct valency, and molecule stability is the fraction of generated molecules in which all atoms are stable. Validity is evaluated using RDKit by checking whether the 3D molecular structures can be successfully parsed into SMILES format. Uniqueness is determined as the ratio of distinct molecules among all valid samples, indicating the diversity of generated molecules. \textcolor{black}{V*U means the ratio of valid and unique molecules.}

\subsection{Unconditional Molecular Generation} 
Unconditional generation assesses the model's ability to learn the underlying molecular data distribution, aiming to generate chemically valid and structurally diverse molecules. 
The results, summarized in Table \ref{exp:main_gen}, show that our method significantly outperforms the baselines across both quality and diversity metrics for the generated molecules. Of particular note is the substantial improvement in the stability of the generated molecules, indicating that our method better satisfies chemical constraints. This validates our hypothesis that our low-noise dynamic is well-suited for generating molecular data.

Additionally, our approach can achieve superior results with significantly fewer generative steps. A detailed comparison, presented in Figure \ref{fig:efficiency}, demonstrates that SLDM achieves 100$\times$ faster sampling than the baseline EDM and around 70$\times$ speedup compared to GeoBFN and EquiFM. \textcolor{black}{Since the methods use the same network architecture, NFE can reflect practical sampling time.} It is important to note that EquiFM utilizes a variety of advanced and efficient ODE solvers, and the results we present correspond to the best performance reported in their paper for different step sizes. \textcolor{black}{Besides, EDM and EDM-Bridge results are from the EDM-Bridge paper while GeoBFN results are reproduced from the official codebase.} This observation underscores the advantages of our straight-line diffusion process: while sophisticated solvers play an important role, the diffusion process design may offer more fundamental improvements. Furthermore, our handcrafted sampling strategies can be seamlessly combined with modern sampling techniques, such as optimal time discretization and advanced solvers, which we leave for future exploration.

  \subsection{Conditional Molecule Generation} 
    Conditional generation evaluates the model's capability to produce molecules with desired properties. Following the baseline approaches, we incorporate property values as additional inputs during training and sample them from a prior distribution during inference. 

    The results in Table \ref{exp:cond_gen} demonstrate that our method consistently outperforms baseline models across all metrics. Moreover, as illustrated in Figure \ref{fig:cond efficiency}, our approach achieves a 20-fold acceleration over previous state-of-the-art methods. This result showcases its potential for application in a wide range of controllable generation scenarios.

\subsection{Ablation Study}\label{sec:ablation unigem}
Previous approaches differ in their methods for atom type generation. EDM represents atom types as one-hot vectors, generating them simultaneously with coordinates through diffusion. In contrast, GeoBFN treats atom types as atomic numbers and generates them by the BFN algorithm for discretized data, which incorporates a binning technique to convert continuous probabilities into discrete probabilities.
UniGEM, on the other hand, generates only coordinates via diffusion and predicts atom types based on the generated coordinates. We adopt the UniGEM framework due to its superior performance.

To compare generative methods without the influence of atom type generation differences, we integrated EDM and GeoBFN coordinate generation algorithms into the UniGEM framework and compared them with our approach. The results, shown in Table \ref{exp:+unigem} with sampling steps $T=50$ and $T=30$, demonstrate that our method exhibits a clear advantage in smaller steps, confirming the efficiency benefits of our generative algorithm. A stability comparison w.r.t various sampling steps is provided in Figure \ref{fig:efficiency-Unigem}.
    Besides, we also conduct an extensive ablation study about temperature control in Appendix \ref{sec:app ablation for temp}.
\begin{figure}[t]
  \begin{minipage}[t]{0.48\textwidth}      
          \begin{table}[H]
        \caption{Conditional generation on QM9 dataset, evaluated by MAE($\downarrow$) between the property condition and the properties of generated molecules predicted by a pretrained EGNN classifier. SLDM uses $T=1000$.}
        \label{exp:cond_gen}
        \small
        \centering
        \scalebox{0.93}{
        \setlength{\tabcolsep}{2.5pt}
        \begin{tabular}{lccccccc}
        \\ \toprule
        Property & $\alpha$ & $\Delta \epsilon$ & $\epsilon^{\text{HOMO}}$ & $\epsilon^{\text{LUMO}}$ & $\mu$  & $C_{v}$ \\
        
        Units & $\text{Bohr}^{3}$ & $\text{meV}$ & $\text{meV}$ & $\text{meV}$ & $D$ &  $\frac{\text{cal}}{\text{mol K}}$ \\
        \midrule
        Data & 0.10 & 64 & 39 & 36 & 0.043 &  0.040 \\
        \hline
        Random & 9.01 & 1470 & 645 & 1457 & 1.616 &  6.857 \\
        
        $N_{\text{atoms}}$ & 3.86 & 866 & 426 & 813 & 1.053 &  1.971 \\
        
        EDM & 2.76 & 655 & 356 & 584 & 1.111  & 1.101 \\
        
        GeoLDM & 2.37 & 587 & 340 & 522 & 1.108 & 1.025 \\
        
        GeoBFN & 2.34 & 577 & 328 & 516 & 0.998 &  0.949 \\
        \hline
        SLDM & \textbf{1.46} &\textbf{440} &\textbf{320} & \textbf{348}  &\textbf{0.797} &\textbf{0.745}\\
        \bottomrule
        \end{tabular}}
        \end{table}
  \end{minipage}
  \hfill
  \begin{minipage}[t]{0.5\textwidth}
    \begin{table}[H]
        \caption{Comparison of generative methods within the UniGEM framework\textcolor{black}{, marked by $_U$, }for unconditional generation on the QM9 dataset using small sampling steps $T$. Higher values indicate better performance.}
        \label{exp:+unigem}
        \small
        \scalebox{0.9}{
        \setlength{\tabcolsep}{1pt}
        \begin{tabular}{clcccc}
        \\ \toprule        
         \ \ T \  \   &Model & Atom sta(\%) & Mol sta(\%) & Valid(\%)   & V*U(\%)  \\
        \midrule
        50&EDM$_U$&98.55&85.73&93.29&91.78 \\
        50&GeoBFN$_U$&98.28&87.16&93.97&91.82 \\     50&SLDM&\textbf{99.30}&\textbf{93.37}&\textbf{96.24}&\textbf{93.63} \\
        \midrule
        30&EDM$_U$&97.58&78.75&89.39&87.96 \\
        30&GeoBFN$_U$&96.74&81.05&90.93&87.47 \\        30&SLDM&\textbf{99.30}&\textbf{93.02}&\textbf{96.20}&\textbf{92.76} \\
        \bottomrule
        \end{tabular}}
        \end{table}
  \end{minipage}
  \end{figure}
  \textcolor{black}{
\subsection{Evaluation on Toy Dataset}
  We provide a toy dataset experiment in Appendix~\ref{sec:app toydata}, demonstrating the superior generative capabilities of our approach in faithfully modeling complex data distributions. These results suggest that SLDM holds promise for generalization to broader application domains.}



\section{Conclusion}
This paper proposes the Straight-Line Diffusion Model (SLDM), a novel generative method that ensures a near-linear diffusion trajectory, effectively reducing truncation error during sampling. The proposed process schedule naturally aligns with the characteristics of point cloud molecular data and effectively balances the generative difficulty. As a result, our method achieves significant improvements in both sampling efficiency and quality, as demonstrated in both unconditional and conditional generation settings, paving the way for large-scale, controllable molecular generation in practice. 

\textcolor{black}{Several challenges remain open for future investigation. On the theoretical side, refining the schedule to fully satisfy the boundary conditions, establishing principled guidelines for selecting the noise level $\sigma$, such as} analyzing its impact on training stability and sampling accuracy, and deriving optimal discretization schemes could strengthen the theoretical foundations of our approach and further accelerate sampling. On the practical side, applying our method to diverse controllable molecular generation tasks and extending it to broader domains presents exciting opportunities for future advancements.

\bibliographystyle{unsrtnat} 

\bibliography{example_paper}

\newpage
\appendix

\section{Supplementary Theoretical Results}\label{sec:app theory}
\subsection{The Derivation of A Unified Formulation of Diffusion Models}\label{secapp:unify formulation}
The stochastic process of the diffusion-based model can be formulated in general as a linear stochastic differential equation (SDE):
\begin{equation}\label{eqapp:sde}
    \mathrm{d}X_t =f(t)X_t \mathrm{d}t+g(t)\mathrm{d}W_t,
\end{equation}  
where $X_t$ characterizes the noise corrupting process of the data in a diffusion model, $f(t)\leq 0$ and $g(t)\geq 0$ are measurable functions defined on the interval $[0,\infty)$, $W_t$ represents a standard Wiener process. The process $X_t$ is a time rescaled Ornstein-Uhlenbeck process whose law converges exponentially fast to the standard Gaussian distribution \cite{chen2023sampling}.

Under the assumption that all the relevant integrals exist, the solution of the above SDE is given by:
\begin{equation}\label{eq:solution}
    X_t=X_0\cdot \exp(\int_0^t f(\xi)\mathrm{d}\xi)+ \int_0^t \exp(\int_s^t f(\xi)\mathrm{d}\xi)g(s)\mathrm{d}W_s, t\in[0,\infty)
\end{equation}
 A quick proof is as follows:
\begin{proof}
By Itô's formula, and applying \eqref{eqapp:sde}:
    \begin{equation}
        \begin{aligned}
            \mathrm{d}(X_t\cdot \exp(-\int_0^t f(\xi)\mathrm{d}\xi))&=(\mathrm{d}X_t-X_tf(t)\mathrm{d}t)\cdot\exp(-\int_0^t f(\xi)\mathrm{d}\xi)\\&=\exp(-\int_0^t f(\xi)\mathrm{d}\xi)\cdot g(t)\mathrm{d}W_t
        \end{aligned}
    \end{equation}
    Next, we integrate both sides and obtain:
    \begin{equation}
        \begin{aligned}
            X_t\cdot \exp(-\int_0^t f(\xi)\mathrm{d}\xi))-X_0=\int_0^t\exp(-\int_0^s f(\xi)\mathrm{d}\xi)\cdot g(s)\mathrm{d}W_s.
        \end{aligned}
    \end{equation}
    Finally, we multiply both sides of the equation by $\exp(\int_0^t f(\xi)\mathrm{d}\xi))$ and rearrange the terms to obtain \eqref{eq:solution}.
\end{proof} 
By the property that the stochastic integral with respect to a Wiener process, $X_t$ is a normal random variable. The mean and variance of $X_t$ are calculated as follows:
\begin{equation}
    \E[X_t]= X_0\cdot exp(\int_0^t f(\xi)\d\xi)
\end{equation}
\begin{equation}
        \mathbb{COV}[X_t]=E[(\int_0^t exp(\int_s^t f(\xi)\d\xi)g(s)\d W_s)^2]\mI=\int_0^t (exp(\int_s^t f(\xi)d\xi)g(s))^2\d s\mI.
\end{equation}
The covariance computation uses the Itô isometry property.

To fit the process $X_t$ to the general form 
\begin{equation}\label{eqapp:general form}
    x_t=\mu(t)x_0+\sigma(t)\epsilon, \epsilon\sim \mathcal{N}(\vzero,\mI),
\end{equation}
we compare it with our solution in \eqref{eq:solution}, and identify that
\begin{equation}
    \begin{aligned}
        &\mu(t)=exp(\int_0^t f(\xi)\d\xi), 
        &\sigma(t)=\mu(t)\sqrt{\int_0^t g(s)^2/\mu(s)^2\d s}.
    \end{aligned}
\end{equation}
Furthermore, we can express $f(t)$ and $g(t)$ in terms of $\mu(t)$ and $\sigma(t)$:
\begin{equation}
\begin{aligned}    
    &f(t)=\frac{\d(\ln\mu(t))}{\d t}=\frac{\dot{\mu}(t)}{\mu(t)},\\
    &g(t)^2=2\sigma(t)\mu(t)\frac{\d(\sigma(t)/\mu(t))}{\d t}=2\sigma(t)\dot{\sigma}(t)-2\sigma(t)^2\frac{\dot{\mu}(t)}{\mu(t)}.
\end{aligned}
\end{equation}
$\mu(t)$ and $\sigma(t)$ satisfy the initial conditions $\mu(0)=1 $ , $\sigma(0)=0$ and additional conditions:
$\mu(t)\geq 0 $ is monotone non-increasing, $\sigma(t)\geq 0$, $\frac{\d(\sigma(t)/\mu(t))}{\d t}\geq 0$, i.e. $\sigma(t)/\mu(t)$ is monotone non-decreasing.

Substituting the expressions for $f(t)$ and $g(t)$ back into the original SDE \eqref{eqapp:sde}, we obtain:
\begin{equation}\label{eqapp:general sde}
    \mathrm{d}X_t =\frac{\dot{\mu}(t)}{\mu(t)}X_t \mathrm{d}t+\sqrt{2\sigma(t)\dot{\sigma}(t)-2\sigma(t)^2\frac{\dot{\mu}(t)}{\mu(t)}}\mathrm{d}W_t,
\end{equation}
This result is conceptually equivalent to that presented in Appendix B of \citet{karras2022elucidating}. But they adopt a different definition of the schedule compared to \eqref{eqapp:general form}, which leads to a different outcome compared to \eqref{eqapp:general sde}.

\subsection{A Summary of Previous Process Schedules}\label{sec:app summarize}
This section provides a comprehensive overview of several widely adopted diffusion-based models and their corresponding process schedules.

The Denoising Diffusion Probabilistic Model (DDPM) \citep{ho2020denoising} is characterized by a process schedule that exhibits the \textbf{Variance Preserving (VP)} property, which can be expressed as: $\mu(t)^2+\sigma(t)^2=1$.
This formulation ensures the preservation of variance across timesteps, assuming the data has unit variance. In the domain of 3D molecular unconditional generation, the Equivariant Diffusion Model (EDM) \citep{hoogeboom2022equivariant} employs a schedule closely resembling the cosine noise schedule introduced by \citet{nichol2021improved}, albeit with a simplified notation:
\begin{equation}
     \vx_t=(1-t^2)\vx_0+\sqrt{1-(1-t^2)^2}\vepsilon, t\in[0,1],
\end{equation}

The \textbf{Variance Exploding (VE)} schedule, initially proposed in the context of Denoising Score Matching \citep{song2019generativescore}, can also be categorized as a denoising diffusion model with a distinct process schedule \citep{karras2022elucidating, song2021scorebased}, which is given by:
\begin{equation}
     \vx_t=\vx_0+\sqrt{t}\vepsilon, t\in[0,T_{max}].
\end{equation}
By rescaling time, we derive the following expression:
\begin{equation}
     \vx_t=\vx_0+\sqrt{t}\sigma_{max}\vepsilon, t\in[0,1],
\end{equation}
where $\sigma_{max}$ needs to be sufficiently large to ensure that $x_1$ approximates a uniform distribution. For illustrative clarity, we set $\sigma_{max}=10$ for Figure \ref{fig:compare schedule mu sigma} and $\sigma_{max}=20$ for Figure \ref{fig:compare schedule trajectory}.

\textbf{Denoising Diffusion Implicit Model (DDIM)} offers an accelerated sampling process for diffusion models. As proved by \citet{karras2022elucidating}, DDIM employs the following schedule:
\begin{equation}
     \vx_t=\vx_0+t\vepsilon, t\in[0,T_{max}].
\end{equation}
By rescaling time, we derive the following expression:
\begin{equation}
     \vx_t=\vx_0+t\sigma_{max}\vepsilon, t\in[0,1],
\end{equation}
For illustrative clarity, we set $\sigma_{max}=10$ for Figure \ref{fig:compare schedule mu sigma} and Figure \ref{fig:compare schedule trajectory}.

\textbf{Flow Matching (FM)} \citep{lipman2023flow} proposes a linear interpolation between the data distribution and a standard Gaussian, with a neural network learning the corresponding vector field. This noise-adding process can also be viewed as defining a diffusion process schedule, given by:
\begin{equation}
     \vx_t=(1-t)\vx_0+(t+(1-t)\sigma_{min})\vepsilon, t\in[0,1],
\end{equation}
where $\sigma_{min}$ needs to be set sufficiently small to ensure that $\vx_t$ aligns with the data distribution at $t=0$. Besides, smaller values of $\sigma_{\text{min}}$ have been reported to yield better results for FM \citep{tong2023conditionalflowmatching}. Accordingly, we set $\sigma_{\text{min}} = 0.001$ for both Figure \ref{fig:compare schedule mu sigma} and Figure \ref{fig:compare schedule trajectory}.

\textbf{Bayesian Flow Networks (BFN)} \citep{graves2023bayesian} is a generative model based on Bayesian inference that accommodates both continuous and discrete variables. For continuous variables, the data is parameterized by Gaussian distributions. In this scenario, the generative algorithms can be interpreted as a denoising diffusion model \cite{xue2024unifyingbfn} with a process schedule given by:
\begin{equation}
     \vx_t=(1-\sigma_{min}^{2t})\vx_0+\sqrt{(1-\sigma_{min}^{2t})\sigma_{min}^{2t}}\vepsilon, t\in[0,1],
\end{equation}
where $\sigma_{min}$ needs to be set small to satisfy $x_t$ align with the data distribution when $t=0$. Following the settings in \cite{song2023geobfn}, we set $\sigma_{min}=0.001$ for Figure \ref{fig:compare schedule mu sigma} and Figure \ref{fig:compare schedule trajectory}.

\subsection{Connection Between Sampling of Baseline Methods and First-Order ODE Sampling}\label{sec:app ode sample}

In section \ref{sec:unify diffusion}, we analyzed the error of first-order ODE discretization methods and claim that the baseline sampling methods primarily rely on first-order ODE discretization. In this section, we provide further clarification on this matter:
EquiFM \cite{song2024equiFM} directly utilizes ODE-based sampling which includes first-order ODE discretization. For methods like EDM \cite{hoogeboom2022equivariant} and GeoBFN \cite{song2023geobfn}, while they resemble first-order methods due to requiring only a single function evaluation per iteration, they incorporate random sampling. This distinction necessitates an explanation of how their random sampling processes relate to ODEs. Specifically, EDM uses the same sampling method as DDPM, which is proved as a first-order discretization to the
 reverse-time SDE of DDPM in Appendix E of \citet{song2021scorebased}. Similarly, GeoBFN adopts the same sampling method as BFN \cite{graves2023bayesian}, which is proved as a first-order discretization to the
 reverse-time SDE of BFN in Proposition 4.2 in \citet{xue2024unifyingbfn}. Please note that both of the reverse-time SDE can be decomposed as an ODE and langevin dynamics:
 \begin{equation}
 \begin{aligned}
     \d\vx_t&=[f(t)\vx_t-g^2(t) \nabla_\vx\log p(\vx_t)]\d t+g(t)\d \vw_t\\
     &=\underbrace{[f(t)\vx_t-\frac{g^2(t)}{2} \nabla_\vx\log p(\vx_t)]\d t}_{\text{reverse time ODE in \eqref{eq:ODE}}}-\underbrace{\frac{g^2(t)}{2} \nabla_\vx\log p(\vx_t)\d t+g(t)\d \vw_t}_{\text{ Langevin dynamics }}     
 \end{aligned}
 \end{equation}
 where $f(t)=\frac{\dot{\mu}(t)}{\mu(t)}$ and $g(t)^2=2\sigma(t)\dot{\sigma}(t)-2\sigma(t)^2\frac{\dot{\mu}(t)}{\mu(t)}$.
 Thus the sampling processes of EDM and GeoBFN can be effectively approximated as a first-order discretization of an ODE augmented by Langevin dynamics. By isolating and analyzing the discretization error of this ODE component, we gain valuable insights into the limitations of methods with first-order discretization, including baseline approaches in molecular generation such as EDM, GeoBFN, and EquiFM.

\subsection{Diffusion Schedule with Straight-line Trajectory}\label{app sec: straight line}
As illustrated in section \ref{sec:straight line}, we aim to reduce the truncation error during sampling by minimizing the second-order derivative of the trajectory. To this end, we first consider a simple case where the initial distribution is a delta distribution $\vx_0\sim \delta_{\va}(\vx)$, $\va\in \R^N$. In this scenario, $\vx_t$ follows a normal distribution $\mathcal{N}(\mu(t)\va,\sigma(t)^2\mI_N)$ according to \eqref{eq:xt def}. The score function is then tractable as
 $\nabla_\vx\log p_t(\vx_t)=-\frac{\vx-\mu(t)\va}{\sigma(t)^2}$. Substituting this into \eqref{eq:ODE} gives: 
\begin{equation}
\begin{aligned}    
    \frac{d\vx}{dt} = \frac{\dot{\sigma}(t)}{\sigma(t)}\vx+\dot{\mu}(t)\va-\frac{\dot{\sigma}(t)}{\sigma(t)}\mu(t)\va 
\end{aligned}
\end{equation}
We aim to keep $\frac{d\vx}{dt}$ constant, which requires $\frac{\dot{\sigma}(t)}{\sigma(t)}=0$ and $\dot{\mu}(t)$ to be constant. This implies that $\sigma(t)$ must be constant, and $\mu(t)$ must be a linear function of $t$. 

Given the boundary condition that $x_{t=0}$ approximates data distribution and $x_{t=T} $ approximates a known distribution, the only feasible choice is to set $\sigma(t)$ to a constant $\mu(t)=1-t/T$. This satisfy all the schedule constraints noted under \eqref{eq:sde}, except $\sigma(0)=0$, which is approximately satisfied by choosing $\sigma$ to be a small value. In practice, we use a value for $\sigma$ that is two orders of magnitude smaller than the data scale, yielding good results. We can also rescale the time as in $[0,1]$, leading to the following schedule for the diffusion process:
\begin{equation}\label{appeq:str schedule}
    \vx_t=(1-t)\vx_0+\sigma\vepsilon, t\in[0,1],
\end{equation}
which ensures a straight-line trajectory under a special data distribution assumption.

Next, we extend the result into general data distribution in the following theorem.
\begin{theorem}\label{thm:app linear_trajectory}
    For a general data distribution and our process schedule $\vx_t=(1-t)\vx_0+\sigma\vepsilon$, $t\in[0,1]$, the following inequality holds for each data dimension $i$:
    \begin{equation}
        P(|\frac{d\vx_t^{(i)}}{dt}+\frac{\vx_t^{(i)}}{1-t}|\geq \delta)\leq\frac{\sigma^2}{\delta^2(1-t)^2}.
    \end{equation}
As $\sigma\to 0$,  the term $\frac{d\vx_t}{dt}+\frac{\vx_t}{1-t}$ converges to zero in probability. More specifically, the solution to the equation $\frac{d\vx_t}{dt}+\frac{\vx_t}{1-t}=0$ is that $\frac{\vx_t}{1-t}$ becomes a constant, which corresponds to a linear trajectory.      
\end{theorem}
In practice, we set $\sigma$ to be small and when \( t \ll 1 - \sigma \), the trajectory is approximately linear.
The proof of the theorem needs two following lemmas.
\begin{lemma}\label{lemma:ode_description}
    For a general data distribution $f(\vx_0)$ and a general diffusion process with the schedule $\vx_t=\mu(t)\vx_0+\sigma(t)\vepsilon$, its ODE description in \eqref{eq:ODE} can be rewritten as 
    \begin{equation}\label{eqapp:ode expectation} 
        \frac{\d\vx_t}{\d t}= \dot{\mu}(t)\E[\vx_0|\vx_t] + \frac{\dot{\sigma}(t)}{\sigma(t)}(\vx_t-\mu(t)\E[\vx_0|\vx_t]).
    \end{equation}
\end{lemma}
\begin{proof}[Proof of Lemma \ref{lemma:ode_description}]
For the general schedule, we have $p_t(\mathbf{x}_t | \mathbf{x}_0) \sim \mathcal{N}(\mu(t)\mathbf{x}_0, \sigma(t)^2 \mathbf{I})$.
We start by considering the score function
\begin{equation}\label{eq: score and expectation}
    \begin{aligned}
        \nabla_{\vx}\log p_t(\vx_t)&=\frac{\int f(\vx_0) \nabla_{\vx} p(\vx_t|\vx_0)  d\vx_0}{p(\vx_t)}  \\
        &= \frac{\int f(\vx_0) p(\vx_t|\vx_0) (-\frac{\vx_t-\mu(t) \vx_0}{\sigma(t)^2})  d\vx_0}{p(\vx_t)}  \\ 
        &= -\frac{\vx_t}{\sigma(t)^2}+\frac{\mu(t) }{\sigma(t)^2}\E[\vx_0|\vx_t].
    \end{aligned}
\end{equation}
Substituting this into the ODE form of the diffusion process, we have:
\begin{equation}
\begin{aligned}   
    \frac{\d\vx_t}{\d t} &=\frac{\dot{\mu}(t)}{\mu(t)}\vx_t-\left(\sigma(t)\dot{\sigma}(t)-\sigma(t)^2\frac{\dot{\mu}(t)}{\mu(t)}\right) (-\frac{\vx_t}{\sigma(t)^2}+\frac{\mu(t) }{\sigma(t)^2}\E[\vx_0|\vx_t])\\
    &=  \dot{\mu}(t)\E[\vx_0|\vx_t] + \frac{\dot{\sigma}(t)}{\sigma(t)}(\vx_t-\mu(t)\E[\vx_0|\vx_t]).
\end{aligned}
\end{equation}
\end{proof}

\begin{lemma}\label{lemma:expectation_and_variance}
   Let $\va$ be a random variable that satisfies $p(\va|\vx_0)\sim\mathcal{N}( \vx_0,\sigma(t)^2/\mu(t) ^2\mI)$. Define $\vy = \E[\vx_0|\va]-\va$. Then, the following properties hold:
   \begin{enumerate}
       \item  $\E [\vy]=\mathbf{0}$,
       \item  $\mathrm{Var} [y^{(i)}]\leq\sigma(t)^2/\mu(t) ^2$, $i=1,...,N$, where $y^{(i)}$ is the $i^{th}$ component of $\vy$. 
   \end{enumerate} 
 
\end{lemma}
\begin{proof}[Proof of Lemma \ref{lemma:expectation_and_variance}]
The expectation of \(\vy\) is given by:
    \begin{equation}
        \E \vy= \E[ \E[\vx_0|\va]-\va]= \E_a\E_{\vx_0|\va}[\vx_0-\va]=  \E_{\vx_0}\E_{\va|\vx_0}[\vx_0-\va]=0
    \end{equation}
    For the \(i\)-th component \(y^{(i)}\), we compute the variance:
    \begin{equation}
    \begin{aligned}
         \mathrm{Var} [y^{(i)}]&=\E [(y^{(i)})^2]= \E[ (\E_{x_{0}|a}[x_0^{(i)}-a^{(i)}])^2]\\&\leq \E[ \E_{x_{0}|a}[(x_0^{(i)}-a^{(i)})^2]]=  \E_{x_{0}}\E_{a|x_{0}}[(x_0^{(i)}-a^{(i)})^2]=\sigma(t)^2/\mu(t)^2
    \end{aligned}       
    \end{equation}
    
The inequality follows from Jensen's inequality.
\end{proof}

\begin{proof}[Proof of \ref{thm:app linear_trajectory}]
For our specific schedule $\sigma(t)=\sigma$ and $\mu(t)=1-t$, \eqref{eqapp:ode expectation} in lemma \ref{lemma:ode_description} reduces to: 
\begin{equation}
  \begin{aligned}  
    \frac{\d\vx_t}{\d t} =-\E[\vx_0|\vx_t] .
\end{aligned}
\end{equation}
For simplicity of notation, we define $\mathbf{a} = \mathbf{x}_t / \mu(t)$, satisfying $p(\mathbf{a} | \mathbf{x}_0) \sim \mathcal{N}(\mathbf{x}_0, \sigma(t)^2 / \mu(t)^2 \mathbf{I})$. By applying lemma \ref{lemma:expectation_and_variance}, the defined 
$\vy = \E[\vx_0|\va]-\va$ satisfies
 $\E [\vy]=\mathbf{0}$ and  $\mathrm{Var} [y^{(i)}]\leq\sigma(t)^2/\mu(t) ^2=\frac{\sigma^2}{(1-t)^2}$, for all dimension $i$.
       
Applying Chebyshev's inequality, we get:
\begin{equation}
    P(|y^{(i)}|\geq \delta)\leq\frac{\mathrm{Var}[y^{(i)}]}{\delta^2}\leq \frac{\sigma^2}{\delta^2(1-t)^2}.
\end{equation}
    Thus, we have the inequality:
    \begin{equation}
        P(|\frac{d\vx_t^{(i)}}{dt}-(-\vx_t^{(i)}/(1-t))|\geq \delta)\leq\frac{\sigma^2}{\delta^2(1-t)^2}.
    \end{equation}

\end{proof}

\subsection{Supplementary Proof for Sampling}\label{secapp:unbiased sampling}
\begin{lemma}[Conditional Expectation Minimizes MSE]\label{lemma: conditional exp}
    The conditional expectation $\E[\vx_{t-\Delta t}|\vx_t]$ is the estimator of $\vx_{t-\Delta t}$ given $\vx_t$ that minimizes the mean squared error (MSE). That is, for any estimator $h(\vx_t)$, the following inequality holds:
    \begin{equation}\label{eqapp:lemma conditional exp}
        \begin{aligned}
            \E[(\vx_{t-\Delta t}-h(\vx_t))^2]\geq \E[ (\vx_{t-\Delta t}-\E[\vx_{t-\Delta t}|\vx_t])^2].
        \end{aligned}
    \end{equation}
\end{lemma}
\begin{proof}
We begin by expanding the conditional squared error term. By the properties of conditional expectation, the cross-term vanishes:
    \begin{equation}
        \begin{aligned}
            &\E[(\vx_{t-\Delta t}-h(\vx_t))^2|\vx_t]=\E[ (\vx_{t-\Delta t}-\E[\vx_{t-\Delta t}|\vx_t]+\E[\vx_{t-\Delta t}|\vx_t]-h(\vx_t))^2|\vx_t]    \\
            &=\E[ (\vx_{t-\Delta t}-\E[\vx_{t-\Delta t}|\vx_t])^2|\vx_t]+\E[ (\vx_{t-\Delta t}-\E[\vx_{t-\Delta t}|\vx_t])(\E[\vx_{t-\Delta t}|\vx_t]-h(\vx_t))|\vx_t]\\& \quad+\E[ (\E[\vx_{t-\Delta t}|\vx_t]-h(\vx_t))^2|\vx_t]\\
            &=\E[ (\vx_{t-\Delta t}-\E[\vx_{t-\Delta t}|\vx_t])^2|\vx_t]+\E[ (\E[\vx_{t-\Delta t}|\vx_t]-h(\vx_t))^2|\vx_t]\\
            &\geq \E[ (\vx_{t-\Delta t}-\E[\vx_{t-\Delta t}|\vx_t])^2|\vx_t]
        \end{aligned}
    \end{equation}
    Taking the expectation with respect to $\vx_t$, we get \eqref{eqapp:lemma conditional exp}, which completes the proof.
    
\end{proof}

    \begin{proposition}    
For the straight-line diffusion schedule defined in \eqref{eq:str schedule}, we have 
 \begin{equation}\label{eq:expectation for ours}
         \E[\vx_{t-\Delta t}|\vx_t] = \frac{1-(t-\Delta t)}{1-t}\left(\vx_t+\sigma^2\nabla_{\vx}\log p_t(\vx_t)\right)   
    \end{equation}
\end{proposition} 
\begin{proof}
    From \eqref{eq: score and expectation}, the conditional expectation of $\vx_0$ given $\vx_t$ is:
    \begin{equation}\label{eq:39}
    \E[\vx_0|\vx_t]=  \frac{1}{\mu(t)}\left(\vx_t+\sigma(t)^2\nabla_{\vx}\log p_t(\vx_t)\right)      
    \end{equation}
    For $\vx_{t-\Delta t}$, the conditional expectation is derived as follows:
    \begin{equation}
    \begin{aligned}        
        \E[\vx_{t-\Delta t}|\vx_t]&=\int \vx_{t-\Delta t}p(\vx_{t-\Delta t}|\vx_t)\d \vx_{t-\Delta t}\\
        &=\int \vx_{t-\Delta t}\int p(\vx_{t-\Delta t}|\vx_0)p(\vx_0|\vx_t)\d \vx_0\d \vx_{t-\Delta t}\\
        &=\int \left(\int \vx_{t-\Delta t}p(\vx_{t-\Delta t}|\vx_0) \d \vx_{t-\Delta t}\right) p(\vx_0|\vx_t)\d \vx_0\\
        &= \int \mu(t-\Delta t)\vx_0  p(\vx_0|\vx_t)\d \vx_0\\
        &= \mu(t-\Delta t)\E[\vx_0|\vx_t],
    \end{aligned}
    \end{equation}
    where we use $x_t$ and $x_t-\Delta t$ are independent given $x_0$. 
    
    Substituting $\E[\vx_0|\vx_t]$ from \eqref{eq:39}, we have:
    \begin{equation}
         \E[\vx_{t-\Delta t}|\vx_t] = \frac{\mu(t-\Delta t)}{\mu(t)}\left(\vx_t+\sigma(t)^2\nabla_{\vx}\log p_t(\vx_t)\right)   
    \end{equation}
Adopting the straight-line diffusion schedule defined in \eqref{eq:str schedule}, we produce \eqref{eq:expectation for ours}.
\end{proof}

As proved in \citep{xue2024sa}, there are a family of reverse processes that share the same marginal probability distributions as \eqref{eq:sde} and \eqref{eq:ODE}. When applied to our process schedule and apply Euler-Maruyama method discretization, we obtain the following iterative algorithm: 
\begin{equation}
\begin{aligned}  \label{eq: 41 general sampling} 
    \vx_{t-\Delta t}
   =& \frac{1-t+\Delta t}{1-t}\vx_t+ (1+\beta(t))\frac{\Delta t}{1-t}\sigma^2 \nabla_\vx \log p_t(\vx_t)+\sqrt{2\beta(t)\frac{\Delta t}{1-t}\sigma^2} \vepsilon.
\end{aligned}
\end{equation}
\eqref{eq: 41 general sampling} uses a Gaussian distribution to model the backward probability $p(\vx_{t-\Delta t}|\vx_t)$, whose expectation is $\frac{1-t+\Delta t}{1-t}\vx_t+ (1+\beta(t))\frac{\Delta t}{1-t}\sigma^2 \nabla_\vx \log p_t(\vx_t)$. According to Lemma \ref{lemma: conditional exp}, the optimal iterative step that that minimizes MSE should satisfy $\frac{1-t+\Delta t}{1-t}\vx_t+ (1+\beta(t))\frac{\Delta t}{1-t}\sigma^2 \nabla_\vx \log p_t(\vx_t)=\frac{1-(t-\Delta t)}{1-t}\left(\vx_t+\sigma^2\nabla_{\vx}\log p_t(\vx_t)\right)  $. This result in $\beta(t)=\frac{1-t}{\Delta t}$.

\subsection{Temperature Control}\label{sec app temp}
For Langevin dynamics, the sampling temperature can be controlled by scaling the stochastic term with a constant $\tau$. 
\begin{equation}\label{eq:LD tau}
d\vx =  \frac{1}{2}g(t)^2\nabla_\vx \log p_t(\vx) dt + \tau g(t) \d w^\prime.
\end{equation}
where \(\tau\) is the temperature parameter. Then $\pi(\vx) \propto p(\vx)^{\frac{1}{\tau^2}}$ is the stationary distribution of the process in \eqref{eq:LD tau}, as proved as follows:
\begin{proof}
    The marginal probability density $\pi_t(x)$ evolves according to Fokker-Planck equation
    \begin{equation}
        \frac{\partial \pi_t(\vx)}{\partial t}=-\nabla\cdot\left[\frac{1}{2}g(t)^2\nabla_\vx \log p_t(x)\pi_t(\vx)\right] + \frac{1}{2}\nabla\cdot\nabla\cdot\left[\tau^2 g(t)^2 \pi_t(\vx)\right]
    \end{equation}
    For the stationary distribution, the probability density becomes time-independent, i.e.,  $\frac{\partial \pi_t(x)}{\partial t}=0$. Thus, we solve:
 \begin{equation}
    \nabla\cdot\left[\frac{1}{2}g(t)^2\nabla_\vx \log p_t(\vx)\pi(\vx)\right] = \frac{1}{2}\nabla\cdot\nabla\cdot\left[\tau^2 g(t)^2 \pi(\vx)\right]
    \end{equation}
    We can easily validate that $\pi(\vx) \propto p(\vx)^{\frac{1}{\tau^2}}$ satisfies the stationary equation.
\end{proof}

Thus, higher temperatures (\(\tau \to \infty\)) increase diversity, with \(\pi(\vx)\) approaching a uniform distribution. Conversely, lower temperatures (\(\tau \to 0\)) enhance fidelity, with \(\pi(\vx)\) converging to a \(\delta\)-distribution at the global maximum of \(p(\vx)\). 

\begin{figure}[h]
    \centering
    \includegraphics[width=0.9\linewidth]{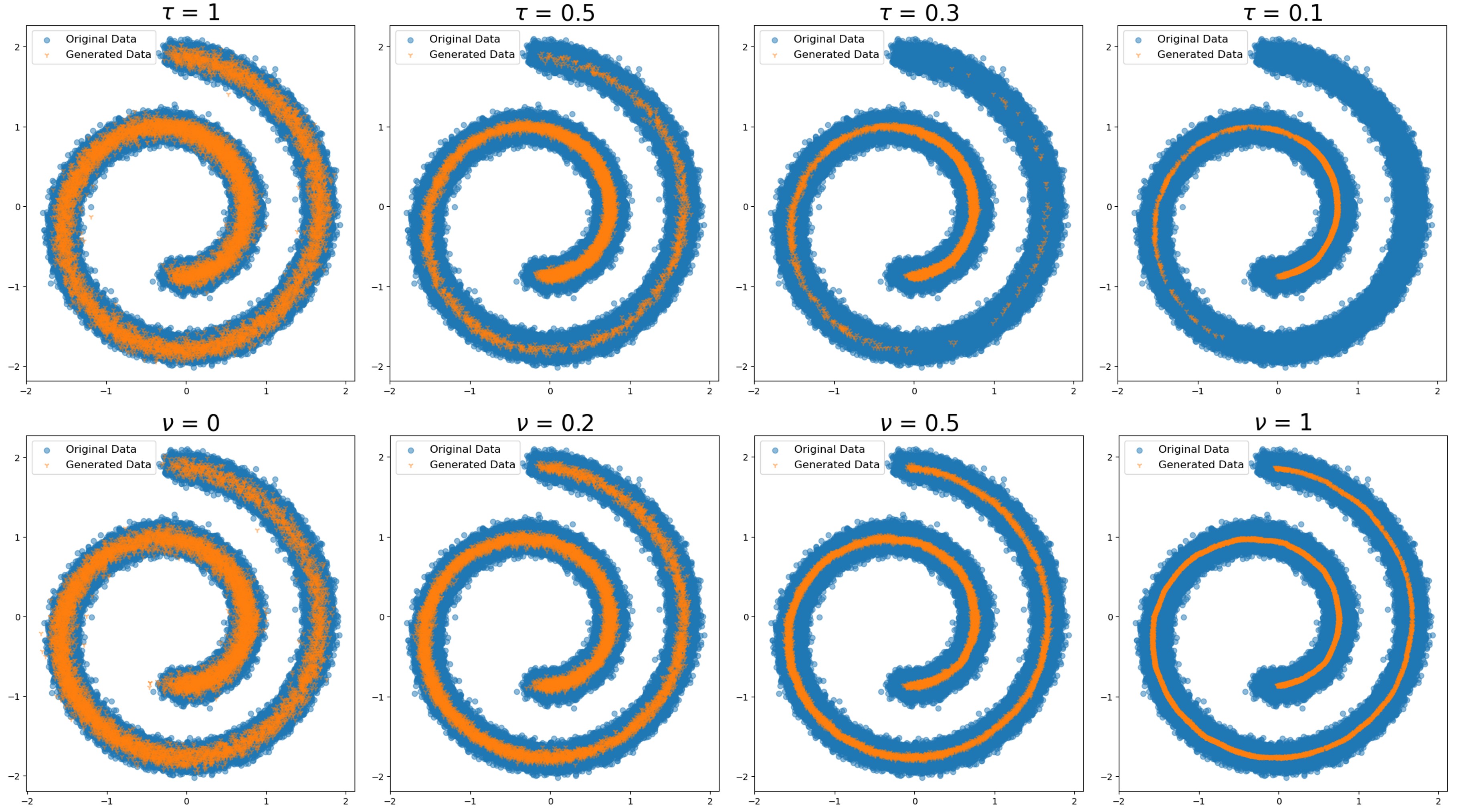}
    \caption{\textbf{a.} Generation results from straight-line diffusion with vanilla temperature control as defined in \eqref{eq:LD tau}.  Training data (blue points) represent 2D Swiss roll coordinates with added Gaussian noise, where the highest density region lies in the interior of the spiral. Diffusion-generated samples (yellow points) exhibit reduced diversity as \(\tau\) decreases but fail to cover all high-density regions, favoring those near the origin. \textbf{b.} Generation results from straight-line diffusion with our annealing temperature control as described in \eqref{eq:LD anneal}. Faster temperature decay (larger \(\nu\)) leads to concentrated samples in high-density regions, successfully covering all local maxima.}
    \label{fig:temp}
\end{figure}

However, prior work has shown that this low-temperature sampling in diffusion models fails to effectively balance diversity and fidelity in image generation, often resulting in blurred and overly smoothed outputs \cite{dhariwal2021diffusion}. We also notice that this low-temperature sampling approach applied to our straight-line diffusion fail to fully cover all high-density regions of the target distribution, as shown in Figure \ref{fig:temp}a.

To address this, we propose an empirically designed time-annealing schedule that introduces higher stochasticity during the initial stages of sampling, allowing for the exploration of distant probability density maxima:

\begin{equation}\label{eq:LD anneal}
    \vx_{t-\Delta t}=\frac{1-t+\Delta t}{1-t}(\vx_t+ \sigma^2\nabla_\vx\log p_t(\vx))+  t^\nu\sqrt{2}\sigma\vepsilon,
\end{equation}

where \(\nu\) controls the decay rate of temperature over time. Larger \(\nu\) enhances molecule stability and enables the model to cover all modes, as shown in Figure \ref{fig:temp}b.
The low-temperature sampling helps mitigate the disturbance caused by noise in the training data, and thereby can improve the quality of the samples. The default value of \(\nu\) for molecular generation is analyzed through an ablation study in Section \ref{sec:app ablation for temp}.

\subsection{Modeling Invariant Probability Density for 3D Coordinate Generation}\label{sec:app equivariance}
In this section, we aim to prove that the probability density of the generated atomic coordinates, as produced by Algorithms \ref{alg:training unigem} and \ref{alg:sampling unigem}, is invariant to both translations and rotations. Formally, we aim to establish that for the Probability Density modeled by our model $p(\vx_0)=p(\vx_0+\vt)$ and $p(\vx_0)=p(\mR\vx_0)$, where $\vt$ is a translation vector and 
$\mR$ is an orthogonal matrix representing a rotation. The proof is in the same spirit of that in \citet{hoogeboom2022equivariant,xu2022geodiff}. \textcolor{black}{The result can be seen as a special case of Theorem 1 and Theorem 2 in \citet{zhou2025physpriors}.}

To ensure translation invariance, the generative process is defined in the quotient space of translations, specifically the zero Center of Mass (CoM) space. This is achieved through two key operations. First, noise is sampled from a CoM-restricted Gaussian distribution $\vepsilon\sim\gN_{\text{CoM}}(\textbf{0}, I_{3M})$, which is sampled by first sampling a standard Gaussian noise vector $\vepsilon'\sim\gN(\textbf{0}, I_{3M})$ and then subtracting its center of mass: $\vepsilon=\vepsilon'-\frac{1}{3M}\sum_{i=1}^{3M}\epsilon'_i$. Second, the network's output at each step is projected into the zero CoM space. These operations ensure that all intermediate coordinates $\vx_t,t=0,\cdots,1$  remain strictly within the zero CoM space throughout the generative process.

To establish rotation invariance, we rely on two key properties. First, we use an equivariant neural network satisfying $\phi^{(x)}(\mR\vx_t,t)=\mR\phi^{(x)}(\vx_t,t)$. Second, we utilize the rotational invariance of the zero mean isotropic Gaussian distributions. This property can also be extended to the CoM-restricted Gaussian distribution, whose probability density function is: $f_{\gN_\text{CoM}}(\vx)=\frac{1}{(2\pi)^{3(M-1)/2}}exp(\frac{1}{2}||\vx||^2)$. Then, we can verify $f_{\gN_\text{CoM}}(\vx)=f_{\gN_\text{CoM}}(\mR\vx)$ for any orthogonal rotation matrix $\mR$. Now we prove the rotation invariance of the generation probability as follows:

At each iterative step of the generative process, we have:
\begin{equation}
    \begin{aligned}
        \vx_{t-\Delta t}\sim \gN_{\text{CoM}}\left(\frac{1-(t-\Delta t)}{1-t}\cdot(\vx_t-\sigma \cdot \phi(\vx_t,t)), 2t^{2\nu}\sigma^2I_{3M}\right)\\
        \end{aligned}
\end{equation}
\begin{equation}
    \begin{aligned}
        p(\mR\vx_{t-\Delta t}|\mR\vx_{t})
        &=f_{\gN_\text{CoM}}\left(\frac{\mR\vx_{t-\Delta t}-\frac{1-(t-\Delta t)}{1-t}\cdot(\mR\vx_t-\sigma \cdot \phi(\mR\vx_t,t))}{ \sqrt{2}t^{\nu}\sigma}\right)\\
        &=f_{\gN_\text{CoM}}\left(\mR\frac{\vx_{t-\Delta t}-\left(\frac{1-(t-\Delta t)}{1-t}\cdot(\vx_t-\sigma \cdot \phi(\vx_t,t))\right)}{ \sqrt{2}t^{\nu}\sigma}\right)\\
        &=f_{\gN_\text{CoM}}\left(\frac{\vx_{t-\Delta t}-\left(\frac{1-(t-\Delta t)}{1-t}\cdot(\vx_t-\sigma \cdot \phi(\vx_t,t))\right)}{ \sqrt{2}t^{\nu}\sigma}\right)=p(\vx_{t-\Delta t}|\vx_{t})
    \end{aligned}
\end{equation}
In the second equality, we apply the equivariance property of the neural network. The third equality follows from the rotational invariance of the isotropic Gaussian distribution.

Additionally, the initial distribution $p(\vx_1)= \gN_{\text{CoM}}(\textbf{0}, \sigma^2 I_{3M})$ is rotation invariant. Combining these facts, we can propagate rotation invariance across the generative process. Thus, for the final generated distribution:
 \begin{equation}
    \begin{aligned}
        p(\mR\vx_0)&=\int\cdots\int\prod_{t=\Delta t}^{1}p(\mR\vx_{t-\Delta t}|\mR\vx_{t}) p(\mR\vx_1) \d \vx_1 \cdots\d \vx_{\Delta t}\\
        &=\int\cdots\int\prod_{t=\Delta t}^{1}p(\vx_{t-\Delta t}|\vx_{t}) p(\vx_1) \d \vx_1 \cdots\d \vx_{\Delta t}= p(\vx_0).
    \end{aligned}
\end{equation}
In the second equality, we use the rotational invariance of both the transition probabilities and the initial distribution. This proves that the final probability density of the generated data is rotation invariant.

\section{Molecular Generation Algorithms}\label{sec:algorithm}
  
\begin{algorithm}
   \caption{Straight-Line Diffusion Training for Molecules (UniGEM)}
   \label{alg:training unigem}
   \KwIn{ 3D molecular data $\vz_0=[\vx_0,\vh_0]$ with $M$ atoms, neural network $\phi$, variance constant $\sigma$, nucleation time $t_n\in[0,1]$}
   \Repeat{$\texttt{converged}$}{
     \hspace{0.2em} Sample $t \sim \frac{1}{2}U([0,t_n])+\frac{1}{2}U([t_n,1])$\\
    Sample $\vepsilon\sim \gN_{\text{CoM}}(\textbf{0}, I_{3M})$ \\
    Compute $\vx_t = (1-t)\vx_0 + \sigma\vepsilon$\\
    Minimize $|| \vepsilon-\phi^{(x)}(\vx_t, t)||^2+ 1_{t\leq t_n} | \vh_0-\phi^{(h)}(\vx_t, t)|$
   }
\end{algorithm}

\begin{algorithm}[h!]
   \caption{Straight-Line Diffusion Sampling for Molecules (UniGEM)}
   \label{alg:sampling unigem}
   \KwIn{Number of atoms $M$, neural network $\phi$, variance constant $\sigma$, nucleation time $t_n$, sampling steps $T$, temperature annealing rate $\nu$}
     Sample $\vx_1 \sim \sigma\cdot\gN_{\text{CoM}}(\textbf{0}, I_{3M})$\\
   \For{$i=T-1$ \textbf{to} $1$}{
    \hspace{0.6em}$t=i/T$, $\Delta t = 1/T$\\
    $\vx_{t-\Delta t}=\frac{1-(t-\Delta t)}{1-t}\cdot(\vx_t-\sigma \cdot \phi(\vx_t,t))$\\
    Projecting $\vx_{t-\Delta t}$ into zero CoM space\\
   \uIf{$i>1$}{
       Sample $\vepsilon\sim \gN_{\text{CoM}}(\textbf{0}, I_{3M})$\\
    $\vx_{t-\Delta t}=\vx_{t-\Delta t}+t^{\nu}\sqrt{2}\cdot\sigma\cdot\vepsilon$}
   \Else{
       $\vh_0= \phi^{(h)}(\vx_0, 0)$
    }
    }
\KwRet{$\vz_0 = [\vx_0, \vh_0]$}
\end{algorithm}

\section{Supplementary Illustrations and Results}\label{sec app: supple result}
\subsection{Sampling Trajectory of Diffusion-based Models}
\begin{figure*}[h!]
	\centering
    \includegraphics[width=1\linewidth]{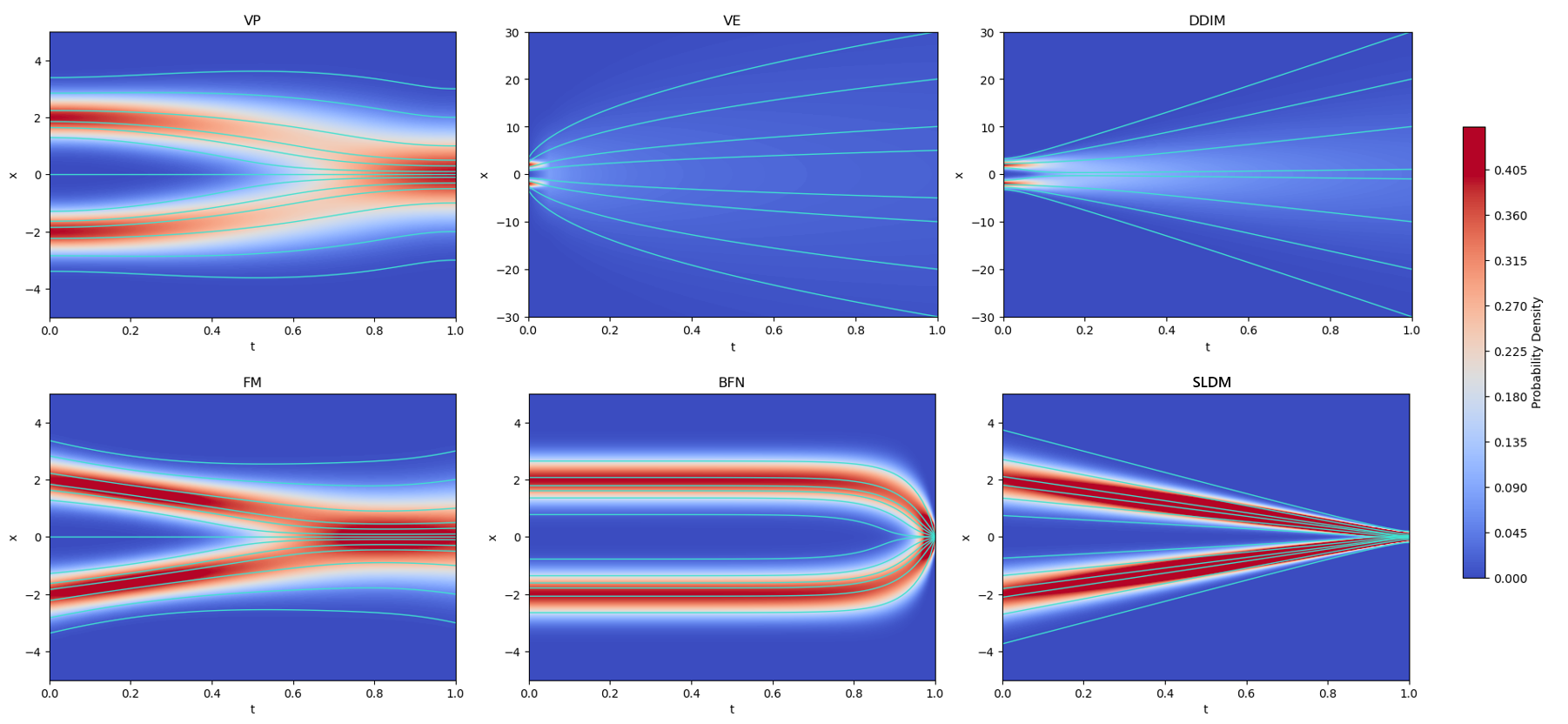}
    \caption{The trajectory of the ODE in \eqref{eq:ODE} of various diffusion-based models. The data distribution is a mixture of Gaussians in one dimension, defined as $x_0\sim 0.5\cdot\gN(2,1/4)+0.5\cdot \gN(-2,1/4)$. The background color indicates the value of probability density $p(x_t)$. \textcolor{black}{During sampling, the timestep $t$ decreases from $1$ to $0$.}}
    \label{fig:compare schedule trajectory}
\end{figure*}

To evaluate whether SLDM exhibits a near-linear trajectory under general data distributions, as suggested by Theorem \ref{thm: linear_trajectory}, we visualize the ODE trajectory and compare it with other diffusion-based models. We consider a scenario where the data distribution is a one-dimensional Gaussian mixture, as this setup offers a tractable score function and serves as a representative example of general distributions. The resulting trajectories are shown in Figure \ref{fig:compare schedule trajectory}.
Our method maintains a trajectory that is consistently closer to a straight line compared to other approaches, leading to smaller truncation errors under first-order numerical discretization. As suggested in Theorem \ref{thm: linear_trajectory}, the trajectory slightly deviates from a straight line in the early stages (i.e., as \(t\) approaches \(1\)), which could increase the sampling error in theory. However, our empirical observations indicate that the sampling process is relatively robust to such errors during these initial steps. This robustness may be attributed to the incorporation of Langevin dynamics in the sampling algorithm, which can mitigate the impact of early-stage inaccuracies. As a result, SLDM achieves an overall lower sampling error.

In comparison, DDIM demonstrates a relatively linear trajectory during the early stages but exhibits significant curvature in later stages, where accurate sampling is more crucial. This could potentially degrade its performance. For the FM method, where the initial distribution is Gaussian and no explicit prior-data joint distribution is predefined as in \citet{tong2023conditionalflowmatching}, it fails to achieve a straight-line trajectory. Other methods also demonstrate an evident curved trajectory. Besides, we can also see from the illustration that the distribution remains nearly static during the later stages of BFN sampling, which aligns with the discussion in section \ref{sec:SNR} and suggests potential inefficiencies of time scheduling.
\subsection{Sampling Efficiency Comparisons under UniGEM Framework}
We complement the ablation study in Section \ref{sec:ablation unigem} by incorporating results with diverse sampling steps. The results demonstrate that while UniGEM enhances the performance of EDM and BFN when sampling step is abundant, these methods still struggle to achieve satisfactory molecular stability in the low sampling step scenario.
\begin{figure*}[h!]
	\centering
    \includegraphics[width=0.6\linewidth]{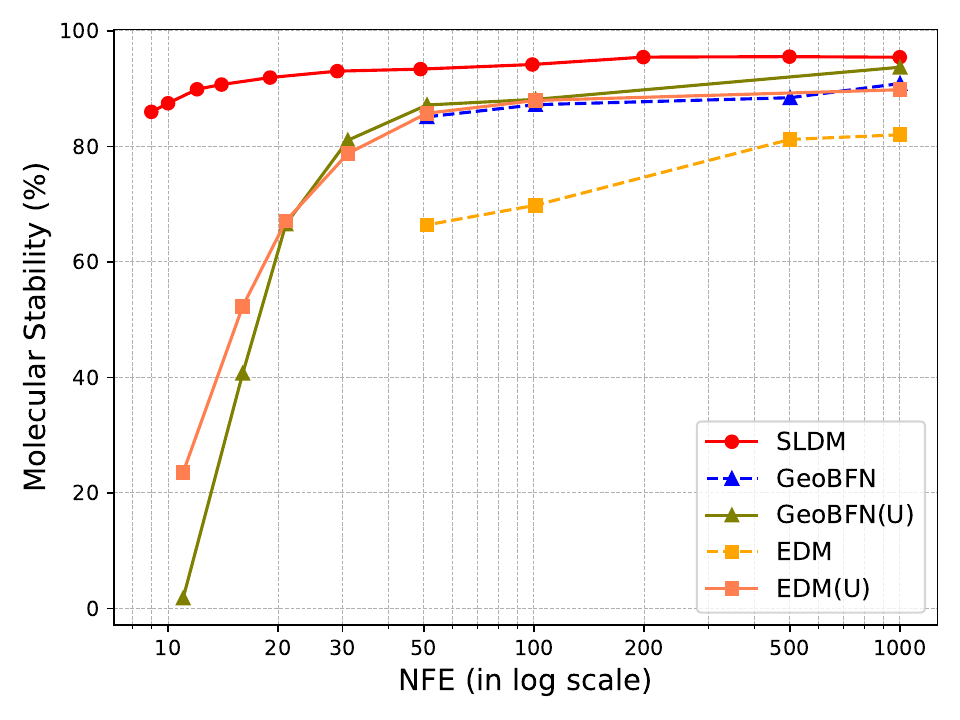}
    \caption{Comparison of molecule stability ($\uparrow$) across diffusion-based molecular generation models with UniGEM framework on QM9 unconditional generation, evaluated with respect to the Number of Function Evaluations (NFE) during the sampling process.}
    \label{fig:efficiency-Unigem}
\end{figure*}

\subsection{Toy Data Results}\label{sec:app toydata}
\begin{figure}[h!]
	\centering
    \includegraphics[width=0.7\linewidth]{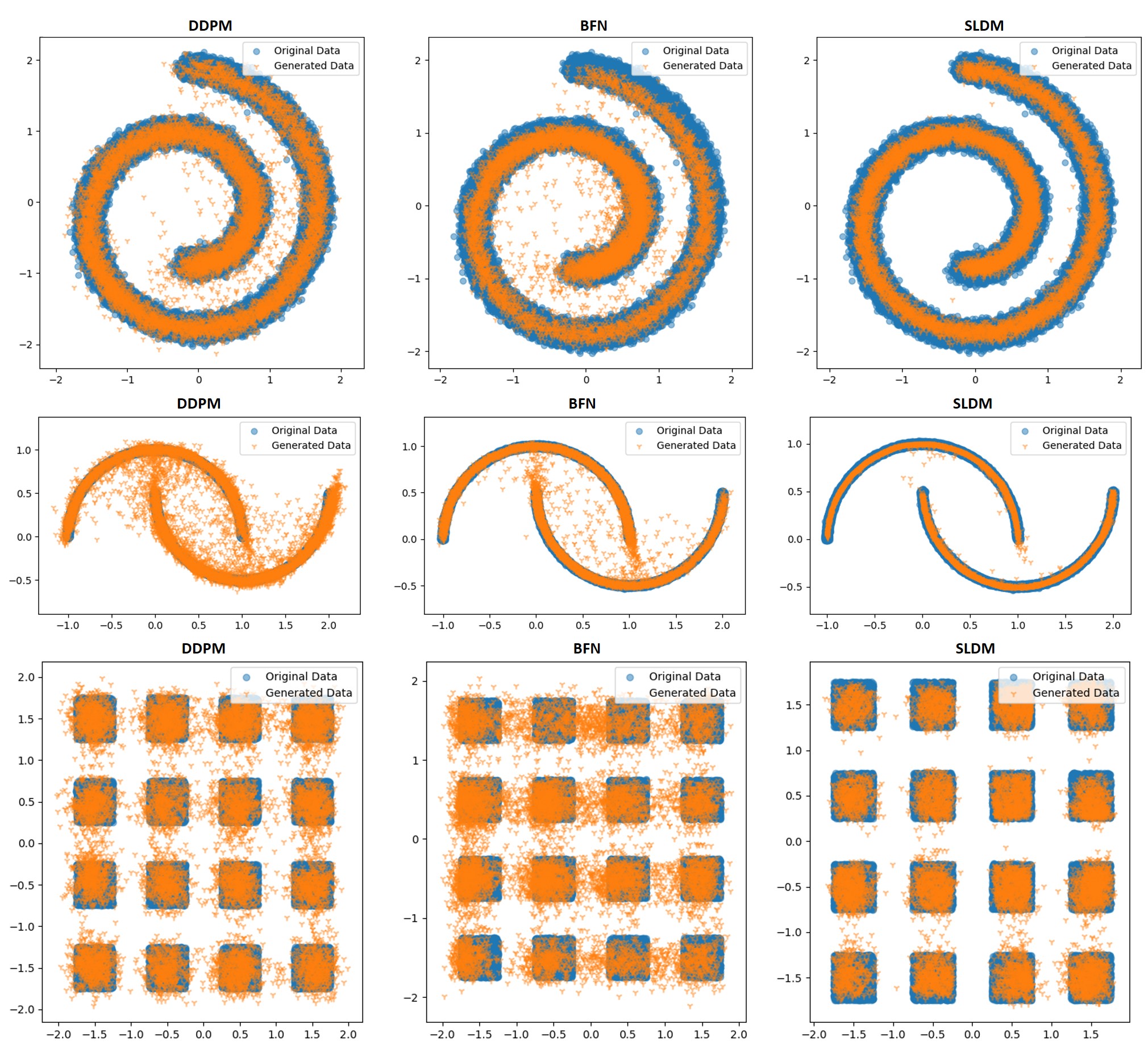}
    \caption{Generative performance comparison of Straight-Line Diffusion, DDPM, and BFN on 2D toy datasets.}
    \label{fig: toy data}
    \vskip -0.1in
\end{figure}

We evaluate our model on several 2D toy data to test its generality. The datasets included swissroll and moons to represent continuous data distributions, as well as chessboard to simulate discretized data distributions. The dataset consists of 100,000 samples. For the moons and swissroll datasets, we performed training with 40 diffusion steps and 100 epochs, and for the chessboard dataset, the training was extended to 600 epochs with 100 diffusion steps to ensure convergence for all models. Other settings are kept the same for all datasets: \textcolor{black}{The model is a 5-layer MLP.} The batch size was set to 2048, and the optimizer used was Adam with a learning rate of 0.001. No temperature control is used during sampling. These settings \textcolor{black}{follow https://github.com/albarji/toy-diffusion/, and} are kept consistent across all generative algorithms to ensure a fair comparison. The results are provided in Figure \ref{fig: toy data}. The data generated by our model show the closest alignment to the original distributions. Specifically, our model produced samples with fewer outliers and achieved good coverage of the data distributions. These findings highlight the superior generative capabilities of our approach in faithfully modeling complex data distribution. 

\subsection{Ablation Study for Temperature Control}\label{sec:app ablation for temp}
Theoretically, a larger temperature annealing rate 
$v$ corresponds to a faster cooling scheme. The results in Table \ref{exp:temp} demonstrate that adjusting the temperature can effectively enhance molecular stability. However, extreme values of 
$v$ may significantly reduce diversity. We selected $v=0.5$ as the default setting for our model, as it achieves the highest 
$U\times V$ score, indicating the optimal balance between diversity and fidelity.
\begin{table}[h]
\caption{Impact of temperature annealing rate evaluated on the QM9 unconditional generation with sampling step $T=50$. Higher values indicate better performance.}
\label{exp:temp}
\centering
\begin{tabular}{ccccc}
\\ \toprule
\multicolumn{1}{l|}{} & Atom sta(\%) & Mol sta(\%) & Valid(\%)   & $U\times V$(\%)  \\
\hline

\multicolumn{1}{l|}{SLDM(v=0)}       & 99.14 & 91.74& 95.40& 
        92.76    \\
        \multicolumn{1}{l|}{SLDM(v=0.5)}       &99.28 & 93.03 & 96.20 & \textbf{93.28} \\
        \multicolumn{1}{l|}{SLDM(v=1)}       &99.27 & 93.46 & 96.08 & 92.42 \\
\multicolumn{1}{l|}{SLDM(v=3)}       &99.37& 94.34& 96.84&         90.08    \\\multicolumn{1}{l|}{SLDM(v=5)}       &99.42 & 94.83 & 97.33 & 86.10 \\
\multicolumn{1}{l|}{SLDM(v=10)}       & 99.53 & 96.06 & 97.98&         75.13     \\

\bottomrule
\end{tabular}
\end{table}

\section{Related Work}
\subsection{An overview of 3D Molecular Generation}\label{sec: related work}
Recent advances in 3D molecular generation can be categorized based on the underlying \textbf{generative algorithms}:
Autoregressive methods generate molecules step by step, progressively connecting atoms or molecular fragments. G-SchNet \citep{gebauer2019symmetry} and G-SphereNet \citep{luo2022autoregressive} are early examples that use this strategy to model 3D molecular structures. Building on these, Symphony \citep{daigavane2024symphony} incorporates higher-degree E(3)-equivariant features and message-passing to improve the modeling of molecular geometries.
Normalizing flow \citep{chen2018continuous} has also been applied to molecule generation by E-NF \citep{garcia2021n} and \citet{kohler2020equivariant}. 
Diffusion-based models have gained prominence for 3D molecular generation recently. Hoogeboom et al. \citep{hoogeboom2022equivariant} introduced the Equivariant Diffusion Model (EDM) to jointly generate atomic coordinates and atom types. Extensions of EDM include EDM-Bridge \citep{wu2022diffusion}, which enhances performance through prior bridges, and GeoLDM \citep{xu2023geometric}, which performs diffusion in a latent space using an autoencoder. EquiFM \citep{song2024equiFM} employs flow matching for efficient molecule generation, and GeoBFN \citep{song2023geobfn} combines Bayesian Flow Networks with distinct generative algorithms tailored for discretized charges and continuous coordinates. \textcolor{black}{END \citep{cornet2024end} proposes a learnable data- and time-dependent noise schedule of the diffusion process, and achieves improved sampling efficiency.}


Several studies offer \textbf{complementary strategies} to generative algorithms.
First, representation-conditioned generation has been explored by MDM \citep{huang2023mdm}, which conditions on VAE representations, and GeoRCG \citep{li2024RCG}, which extends EDM by leveraging pretrained molecular representations.
Second, some studies propose advanced network architectures to enhance molecular generation \citep{hua2024mudiff, huang2023mdm, le2024navigating, irwin2024efficient}\textcolor{black}{\citep{jodo,Twigs}}.
Third, additional input information has been introduced to the molecular generation. For instance, MolDiff \citep{peng2023moldiff} explicitly predicts bonds during generation. MiDi \citep{vignac2023midi}, \textcolor{black}{JODO \citep{jodo},} EQGAT-diff \citep{le2024navigating}, and SemlaFlow \citep{irwin2024efficient} further extend generation to include bonds and formal charges, enriching the molecular inputs. \textcolor{black}{Twigs \citep{Twigs}, on the other hand, integrates additional molecular properties into the diffusion training process, leading to enhanced conditional generation.}

It is worth noting that \textbf{evaluation strategies} for 3D molecular generation can be different across methods. EDM \citep{hoogeboom2022equivariant} employs strict rules for bond definitions based on interatomic distances, implicitly enforcing constraints on bond lengths and steric hindrance. In this framework, the metric "stability" is rigorously defined, requiring correct valency and neutral atomic charges. Our method, along with most diffusion-based approaches \citep{wu2022diffusion, xu2023geometric, song2023geobfn}, adheres to this evaluation standard.
In contrast, another line of studies \citep{vignac2023midi, le2024navigating, irwin2024efficient} infers bonds and formal charges directly from the model, allowing atoms to have non-zero formal charges. Therefore, the bond inference imposes no constraints on bond lengths or steric hindrance. Further, the "stability" metric is defined more loosely, permitting discrepancies between valency and the number of covalent bonds. This relaxed evaluation framework makes it challenging to directly compare these two approaches. We advocate future efforts to establish more equitable evaluation methods to fairly assess the strengths of both paradigms.

\subsection{Related Diffusion-Based Studies and Key Differences with SLDM}\label{sec:discussion Relevant Techniques}
 \textbf{Noise Scheduling} It is important to note that our proposed process schedule differs fundamentally from previous works on noise scheduling \citep{karras2022elucidating, nichol2021improved, le2023eqgat-diff}, time discretization strategies \citep{xue2024accelerating, li2023autodiffusion}, and adaptive step size methods \citep{lu2022dpm-solver}. These approaches can be interpreted as applying a time transformation $\mathcal{T}(\cdot)$ that jointly rescales the diffusion process schedule as $\mu(\mathcal{T}(t))$ and $\sigma(\mathcal{T}(t))$. Notably, such transformations preserve the monotonicity and endpoints of the schedule functions. In contrast, our method decouples $\mu$ and $\sigma$, and fundamentally alters the monotonicity and endpoint of $\sigma(t)$.

 \textbf{Linear $\mathbf{\vmu(t)}$.} Although our method is derived from the diffusion perspective, its process schedule shares some similarities with flow matching (FM) algorithms, such as FM-OT \citep{lipman2023flow} and conditional flow matching (CFM) \citep{tong2023conditionalflowmatching}. Both methods employ a linear $\mu(t)$, and CFM introduces additional low-scale noise. From this perspective, our process schedule can also be interpreted as a variant of CFM. Specifically, our approach employs a prior distribution modeled as a small-scale Gaussian distribution centered at the origin. However, FM and diffusion differ fundamentally in their perspectives: FM models the generative process as an ODE, while diffusion models a stochastic process. This core distinction leads to differences in both the learning targets and sampling methods. Therefore, unlike FM that learns the velocity field and relies on ODE-based sampling, our approaches focus on learning the noise and employ stochastic sampling.
 
 \textbf{Straight Sampling Trajectory.} Moreover, similar to our approach, flow-based methods also aim at flows with straight trajectories. However, these methods rely on predefined prior-data joint distribution to produce straighter paths. This distribution is procured by solving an optimal transport (OT) problem during the training of flow matching \citep{tong2023conditionalflowmatching, song2024equiFM}, and solving such problem is often challenging. When the OT solution is unavailable, achieving straight-line trajectories typically requires additional distillation steps or solving optimization problems, as studied in \textcolor{black}{rectified flow \citep{liu2022rectifiedflow}, progressive distillation \cite{salimans2022progressive}, consistency model \cite{song2023consistency,luo2023latent} and optimal flow
 matching} \citep{kornilov2024optimalflowmatching}. \textcolor{black}{NFDM \citep{bartosh2024neural} proposes a learnable forward process to adapt to align with the reverse process and introduces penalties on the curvature of the reverse process’s trajectories.} In contrast to these existing methods, our approach offers a more straightforward solution for achieving a straight trajectory, by designing a novel diffusion process that minimizes the second-order derivative of the trajectory. 

\section{Implementation Detail} \label{sec:app detail}
The hyperparameter settings for molecular generation are detailed in Table~\ref{tab:hyperparameter}. 
Settings follow UniGEM \cite{feng2024unigem}, with two additional tunable hyperparameters introduced by our generative algorithm: the noise variance $\sigma$ and the temperature annealing rate $\nu$. 

\begin{table}[h]
\centering
\caption{Network and training hyperparameters.}
\label{tab:hyperparameter}
\begin{tabular}{ll}
\toprule
Network Hyperparameters &Value \\
\midrule
Embedding size & 256 for unconditional generation, 192 for conditional generation \\
Layer number & 9 for QM9, 4 for Geom-Drugs \\
Shared layers & 1 \\
\midrule
Training Hyperparameters & Value \\
\midrule
Batch size & 64 for QM9, 128 for Geom-Drugs \\
Train epoch & 3000 for QM9, 32 for Geom-Drugs \\
Learning rate & $1.00 \times 10^{-4}$ \\
Optimizer & Adam \\
Sample steps $T$  & 10 $\sim$ 1000 \\
Nucleation time & 10 \\
Oversampling ratio & 0.5 for each branch\\
Loss weight & 1 for each loss term \\
\midrule
Generative Algorithm Hyperparameters & Value \\
\midrule
Noise Variance $\sigma$ & $0.05$ for unconditional generation, $0.1$ for conditional generation\\
Temperature Annealing Rate $\nu$ & $0.5$ for unconditional generation, $3$ for conditional generation\\
Non-uniform Discretization & False if $T>13$\\
\bottomrule
\end{tabular}
\end{table}

For QM9, it takes approximately 10 days on a single A100 GPU. For GEOM-drugs, it takes approximately 16 days on four A100 GPUs.
For sampling steps greater than 13, the geometric straight-line diffusion use a uniform time discretization like GeoBFN and EDM. However, according to theorem \ref{thm: linear_trajectory}, our trajectory exhibits a larger second-order derivative at the beginning of sampling. Therefore, a more efficient discretization strategy is to use fine-grained discretization for larger $t$ values. We manually set an empirical discretization strategy that yields a 1\% to 10\% improvement in Mol Stable when $T\leq 13$. For sampling steps greater than 13, the impact on the results is less significant ($<1$\%). We leave the exploration of the optimal discretization strategy for straight-line diffusion for future work.

\textcolor{black}{For conditional molecular generation, we follow the baseline approaches \citep{hoogeboom2022equivariant} by incorporating property values as additional inputs during training and sampling them from a prior distribution during inference. The results are measured using a property classifier network trained on the first half of the QM9 dataset, while the remaining portion is used for training the generative model. In addition to generative model baselines, we include two basic baselines: Random and $N_{\text{atoms}}$. The Random baseline involves shuffling property labels in the training data and evaluating the property classifier on the shuffled data. The $N_{\text{atoms}}$ baseline uses the property classifier network to predict the property based solely on the number of atoms in the molecule.}

\end{document}